\documentclass[10pt,twocolumn,letterpaper]{article}
\usepackage{iccv}
\usepackage{times}
\usepackage{epsfig}
\usepackage{graphicx}
\usepackage{amsmath}
\usepackage{amssymb}
\usepackage{array}
\usepackage{multirow}
\usepackage{subcaption}
\usepackage{epstopdf}
\usepackage{xcolor, soul}
\usepackage{booktabs}
\newcolumntype{P}[1]{>{\centering\arraybackslash}p{#1}}
\usepackage[pagebackref=true,breaklinks=true,letterpaper=true,colorlinks,bookmarks=false]{hyperref}

\iccvfinalcopy 

\def\iccvPaperID{10161} 

\ificcvfinal\fi

\begin{document}

\title{HyperReenact: One-Shot Reenactment via Jointly Learning to Refine and Retarget Faces}

\author{\parbox{16cm}{\centering
    {\large Stella Bounareli$^1$,  Christos Tzelepis$^2$,  Vasileios Argyriou$^1$,  Ioannis Patras$^2$, \\ Georgios Tzimiropoulos$^2$ }\\
    {\normalsize
    $^1$ School of Computer Science and Mathematics, Kingston University London\\
    $^2$ School of Electronic Engineering and Computer Science, Queen Mary University of London}}
}

\ificcvfinal\thispagestyle{empty}\fi

\twocolumn[{%
\renewcommand\twocolumn[1][]{#1}%
\vspace{-1em}
\maketitle
    \begin{center}
    \centering 
    \includegraphics[width=0.9\linewidth]{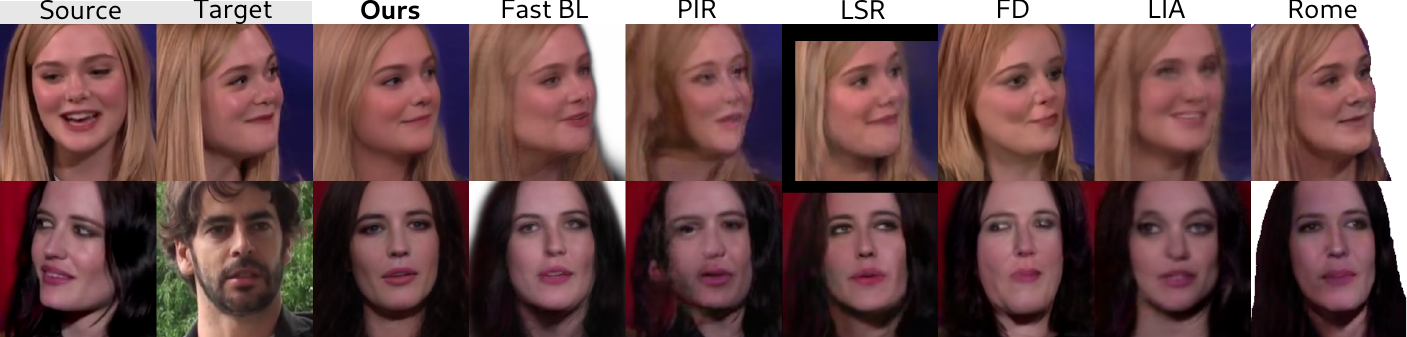}
    \captionof{figure}{The proposed method, named HyperReenact, aims to synthesize realistic talking head sequences of a source identity driven by a target facial pose (i.e., 3D head orientation and facial expression). Our method performs both self and cross-subject reenactment and operates under the one-shot setting (i.e., using a single source frame). We demonstrate that the proposed framework can effectively reenact the source faces without producing significant visual artifacts, even on the challenging conditions of extreme head pose difference between the source and the target images (first row) and on cross-subject reenactment (second row). We compare our method against several state-of-the-art works on neural face reenactment, namely Fast BL~\cite{zakharov2020fast}, PIR~\cite{ren2021pirenderer}, LSR~\cite{meshry2021learned}, FD~\cite{bounareli2022finding}, LIA~\cite{wang2021latent} and Rome~\cite{khakhulin2022rome}.}
    \label{fig:teaser}
    \end{center}
}]

\maketitle

\begin{abstract}
   In this paper, we present our method for neural face reenactment, called HyperReenact, that aims to generate realistic talking head images of a source identity, driven by a target facial pose. Existing state-of-the-art face reenactment methods train controllable generative models that learn to synthesize realistic facial images, yet producing reenacted faces that are prone to significant visual artifacts, especially under the challenging condition of extreme head pose changes, or requiring expensive few-shot fine-tuning to better preserve the source identity characteristics. We propose to address these limitations by leveraging the photorealistic generation ability and the disentangled properties of a pretrained StyleGAN2 generator, by first inverting the real images into its latent space and then using a hypernetwork to perform: (i) refinement of the source identity characteristics and (ii) facial pose re-targeting, eliminating this way the dependence on external editing methods that typically produce artifacts. Our method operates under the one-shot setting (i.e., using a single source frame) and allows for cross-subject reenactment, without requiring any subject-specific fine-tuning. We compare our method both quantitatively and qualitatively against several state-of-the-art techniques on the standard benchmarks of VoxCeleb1 and VoxCeleb2, demonstrating the superiority of our approach in producing artifact-free images, exhibiting remarkable robustness even under extreme head pose changes. We make the code and the pretrained models publicly available at: \url{https://github.com/StelaBou/HyperReenact}.
\end{abstract}

\section{Introduction}\label{sec:intro}
    
    The recent developments in deep learning and generative models~\cite{karras2019style, karras2020analyzing} have led to remarkable progress in the field of facial image synthesis and editing. Among the tasks that have drawn benefit from this progress is neural face reenactment, that aims to synthesize photorealistic head avatars. Specifically, given a source and a target image, the goal of face reenactment is to generate a new image that conveys the identity characteristics of the source face and the facial pose (defined as the 3D head orientation and facial expression) of the target face. The key objectives of this task are three-fold: (i) creating realistic facial images that resemble the real ones, (ii) preserving the source identity characteristics, such as the facial shape, and (iii) faithfully transferring the target facial pose. This technology is an essential component within numerous applications of augmented and virtual reality, as well as arts and entertainment industries. However, despite the recent advancements, most of the existing reenactment methods fail in producing realistic facial images in the \textit{one-shot} setting (i.e., using a single source frame) or under \textit{extreme head pose} movements (i.e., large differences in the head pose of the source and the target).

    The majority of the state-of-the-art methods in neural face reenactment (e.g.,~\cite{zakharov2020fast, meshry2021learned, khakhulin2022rome}) train controllable models that learn to synthesize realistic images. However, these methods are prone to severe visual artifacts, especially when the source and the target faces have large head pose differences. Most of these methods rely on paired data training (i.e., images of the same identity), limiting their applicability in cross-subject reenactment. Several methods~\cite{zakharov2019few, burkov2020neural, meshry2021learned, hsu2022dual} require expensive few-shot fine-tuning (i.e., using multiple different views of the source face) in order to faithfully preserve the source identity and appearance. Another line of research leverages the exceptional generation ability of pretrained generative adversarial networks (GANs)~\cite{bounareli2022finding,bounareli2022StyleMask,yin2022styleheat}, achieving to effectively disentangle the identity from the facial pose. However, these works rely on external real image GAN inversion methods and, thus, are bounded by their limitations, such as poor identity reconstruction and image editability~\cite{tov2021designing}.
    
    In this paper, we draw inspiration from recent works that combine a GAN generator with a hypernetwork~\cite{hypernets} for real image inversion, namely HyperStyle~\cite{alaluf2022hyperstyle} and HyperInverter~\cite{dinh2022hyperinverter}. These methods use a hypernetwork~\cite{hypernets}, conditioned on features derived from the original image and its initial inversion, that learns to modify the weights of the generator to obtain improved image reconstruction quality. Subsequently, one can perform semantic editing of the refined image in the latent space and produce the edited image using the updated generator weights. In spite of their high-quality reconstruction results, HyperStyle~\cite{alaluf2022hyperstyle} and HyperInverter~\cite{dinh2022hyperinverter} fail upon applying global editings on the inverted images and, consequently, they are not practically applicable to neural face reenactment.
    
    In order to address the limitations of state-of-the-art works, we tackle the neural face reenactment task by leveraging the photorealistic image generation and the disentangled properties of a pretrained StyleGAN2~\cite{karras2020analyzing}, along with a hypernetwork~\cite{hypernets}. We present a novel method that performs both faithful identity reconstruction and effective facial image editing by learning to update the weights of a StyleGAN2 generator using a hypernetwork approach. Specifically, our model effectively combines the appearance features of a source image and the facial pose features of a target image to create new facial images that preserve the source identity and convey the target facial pose.
    
    Overall, the main contributions of this paper can be summarized as follows:
    \begin{enumerate}
        \item We present a novel framework for face reenactment that leverages the adaptive nature of hypernetworks~\cite{hypernets} to alter the weights of a powerful StyleGAN2~\cite{karras2020analyzing}, so as to perform both: (i) refinement of the source identity details and (ii) reenactment of the source face in the target facial pose. To the best of our knowledge, we are the first to show the effectiveness of merging the steps of inversion refinement and facial pose editing for robust and realistic face reenactment.
        \item We demonstrate that our method is able to successfully operate under one-shot settings (i.e. using \textit{a single} source frame) without requiring any fine-tuning, preserving the source identity characteristics on both self and cross-subject reenactment scenarios. This holds true even in challenging cases where the source face is partially self-occluded (i.e., in partial facial views due to highly non-frontal head poses). 
        \item We show that our method achieves state-of-the-art results even on \textit{extreme} head pose variations, generating artifact-free images and exhibiting remarkable robustness to large head pose shifts.
        \item We conduct experiments on the standard benchmarks of VoxCeleb1 and VoxCeleb2~\cite{Nagrani17,Chung18b}, performing qualitative and quantitative comparisons with existing state-of-the-art reenactment techniques. We show that our proposed method achieves compelling results both on identity preservation and facial pose transfer.
    \end{enumerate}

\section{Related work}\label{sec:related_work}
    
    \noindent \textbf{Facial image editing} Several recent methods~\cite{shen2020interfacegan, tewari2020stylerig, tzelepis2021warpedganspace,barattin2023attribute,oldfield2021tensor,oldfield2023panda,yang2021discovering,abdal2021styleflow,patashnik2021styleclip} leverage the remarkable ability of modern pretrained GAN models (e.g., StyleGAN2~\cite{karras2020analyzing}) in producing photorealistic facial images in order to edit various facial attributes, such as head pose, facial expressions or hair style. The key idea of such methods lies in learning to manipulate the latent representations (in $\mathcal{W}$, $\mathcal{W}^+$ or $\mathcal{S}$ latent space) of a StyleGAN2 generator in order to generate meaningful editings of different semantics on the synthetic images. The existing unsupervised methods~\cite{voynov2020unsupervised,tzelepis2021warpedganspace} are able to find disentangled directions that edit different semantics on the facial images, albeit without providing any controllability on the manipulation of the images. In order to allow for explicit control over the image editing, several approaches~\cite{abdal2021styleflow,bounareli2022StyleMask,tewari2020stylerig,patashnik2021styleclip,tzelepis2022contraclip,yang2023just} rely on external supervision from pretrained models, such as 3D Morphable Models (3DMM)~\cite{blanz1999morphable}, or vision-language models~\cite{radford2021learning}. Despite their effectiveness in editing synthetic images, such methods fail to manipulate effectively real images -- i.e., having to perform editing in the non-native latent code provided by an external GAN inversion method, as discussed below.
    
    \noindent \textbf{Real image inversion:} GAN inversion methods~\cite{xia2022gan,tov2021designing,vsubrtova2022chunkygan} allow for encoding of real images into the latent space of pretrained generators, which is required at the same time to allow for semantic image editing. The main challenges of real image inversion are to (i) faithfully reconstruct the real images and (ii) enable facial image editing without producing visual artifacts. This is typically referred to as the ``reconstruction-editability'' trade-off. The existing inversion methods mainly focus either on optimization-based approaches, which require expensive iterative optimization for each image, rendering them not-applicable for real-time applications, or on encoder-based architectures. 

    Encoder-based methods~\cite{alaluf2021restyle,richardson2021encoding,tov2021designing,wang2022high,bai2022high} train encoders that learn to predict the latent code that best reconstructs the real image. While providing better semantic editing than optimization-based approaches, encoder-based methods fail in faithfully reconstructing real images (i.e., by missing crucial identity details). In order to coordinate the trade-off between reconstruction quality, editability, and inference time, some recent works~\cite{alaluf2022hyperstyle, dinh2022hyperinverter} propose to optimize a hypernetwork~\cite{hypernets} that learns to update the weights of a pretrained GAN generator so as to refine any missing identity details. HyperStyle~\cite{alaluf2022hyperstyle} first inverts the real images into the latent space of StyleGAN2 ($\mathcal{W}$ or $\mathcal{W}^+$) using a pretrained encoder-based inversion method~\cite{tov2021designing} and then trains a hypernetwork that, given a pair of a real and an initial reconstructed image, predicts an offset $\Delta_{\ell}$ for each layer of the generator. Both HyperStyle~\cite{alaluf2022hyperstyle} and HyperInverter~\cite{dinh2022hyperinverter} lead in high quality reconstructions, albeit suffering from many visual artifacts in the case of head pose editing.

    \noindent \textbf{Neural face reenactment}  A recent line of works focus on learning disentangled representations for the identity and the facial pose using facial landmarks~\cite{zakharov2019few,zakharov2020fast, huang2020learning, ha2020marionette,zhang2020freenet}. However, such methods perform poorly on the challenging task of cross-subject reenactment, since facial landmarks preserve the facial shape and consequently the identity geometry of the target face. In order to mitigate the identity leakage from the target face to the source face, several methods~\cite{koujan2020head2head, ghosh2020gif, tewari2020stylerig,bounareli2022finding,doukas2021headgan,ren2021pirenderer,khakhulin2022rome,Face2Face} leverage the disentangled properties of 3D Morphable Models (3DMM)~\cite{blanz1999morphable, feng2020deca}. Xu et al.~\cite{xu2022designing} propose a unified architecture that learns to perform both face reenactment and swapping. Warping based methods~\cite{wiles2018x2face,siarohin2019first,wang2021one,ren2021pirenderer,doukas2021headgan, zhang2021flow, yin2022styleheat,wang2021latent} learn a motion field between the source and target frames in order to synthesize the reenacted faces. \cite{doukas2021headgan,ren2021pirenderer} propose a two-stage architecture that first generates a warped image using the learned motion field and then refines the warped image to minimize the visual artifacts caused by the warping operation. Despite their realistic results in small pose variations, such methods fail in the more challenging and realistic condition of large head pose variations (i.e., under large differences between the target and the source head pose).

    A more recent line of works~\cite{bounareli2022finding,yin2022styleheat,bounareli2022StyleMask} propose the incorporation of the powerful pretrained StyleGAN2 model. StyleHEAT~\cite{yin2022styleheat} proposes to control the spatial features of the pretrained StyleGAN2 generator using a learned motion field between the source and target frames. However, their method is trained on the HDTF dataset~\cite{zhang2021flow}, i.e., a video dataset with mostly frontal talking head videos, leading to poor reenactment performance on more realistic datasets, such as VoxCeleb~\cite{Nagrani17, Chung18b}, which comprises of a larger distribution on the existing head poses. Bounareli et al.~\cite{bounareli2022finding} propose to fine-tune a StyleGAN2 model (pretrained on FFHQ~\cite{karras2019style}) on VoxCeleb1 dataset~\cite{Nagrani17} and then learn the linear directions that are responsible for controlling the changes in the facial pose. For editing the real images,~\cite{bounareli2022finding} relies on the encoder-based inversion method of~\cite{tov2021designing}, which results in visual artifacts on large head pose variations and requires an additional optimization step~\cite{roich2021pivotal} to refine the missing identity details. Finally, \cite{bounareli2022StyleMask} proposes to disentangle the identity characteristics from the facial pose by leveraging the disentangled properties of StyleGAN2's style space.

    In this work, we also use the StyleGAN2 generator, pretrained on VoxCeleb~\cite{bounareli2022finding}, which allows for better generalization on other existing video datasets~\cite{zhang2021flow,roessler2018faceforensics}. To the best of our knowledge, we are the first to propose the optimization of a hypernetwork~\cite{hypernets} based reenactment module, merging this way the steps of inversion refinement and facial pose editing towards robust and realistic face reenactment.

\section{Proposed method}\label{sec:proposed_method}
    
    \begin{figure*}[t!]
        \centering
        \includegraphics[width=0.95\textwidth]{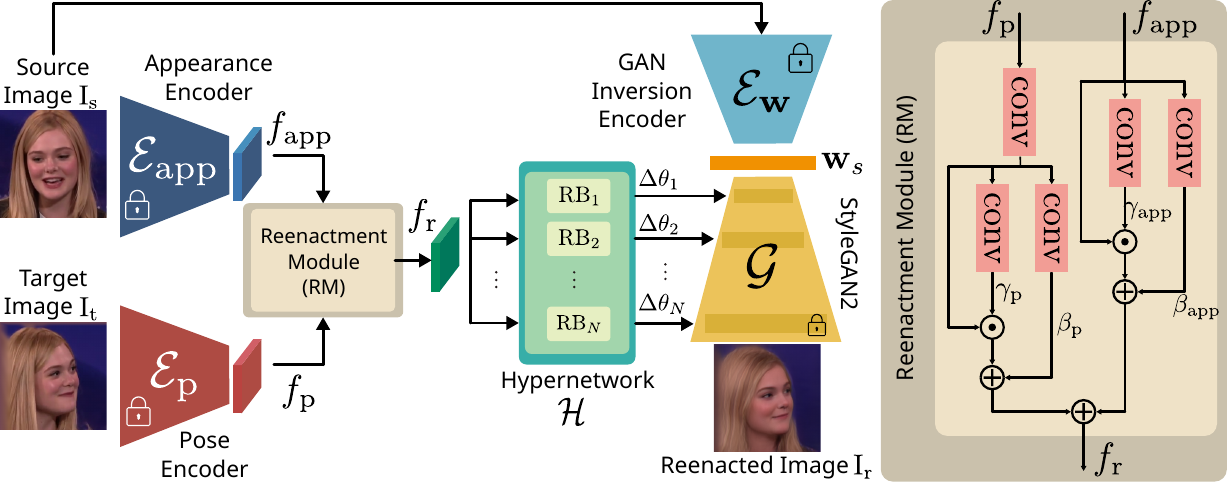}
        \caption{\textbf{HyperReenact network architecture} Given a source ($\mathrm{I}_{\mathrm{s}}$) and a target ($\mathrm{I}_{\mathrm{t}}$) image, we first extract the source appearance features, $f_{\mathrm{app}}$, and the target pose features, $f_{\mathrm{p}}$, using the appearance ($\mathcal{E}_{\mathrm{app}}$) and the pose ($\mathcal{E}_{\mathrm{p}}$) encoders, respectively. The Reenactment Module (RM) learns to effectively fuse these features, producing a feature map $f_{\mathrm{r}}$ that serves as input into each Reenactment Block (RB) of our hypernetwork module $\mathcal{H}$. The predicted offsets, $\Delta\theta$, update the weights of the StyleGAN2 generator $\mathcal{G}$ so that using the inverted latent code $\mathrm{w}_\mathrm{s}$ generates a new image $\mathrm{I}_{\mathrm{r}}$ that conveys the identity characteristics of the source face and the facial pose of the target face. We note that, during training, the encoders $\mathcal{E}_{\mathrm{app}}$, and $\mathcal{E}_{\mathrm{p}}$, along with the generator $\mathcal{G}$ are kept frozen, and we optimize only the Reenactment Module (RM) and the hypernetwork module $\mathcal{H}$.}
        \label{fig:architecture}
    \end{figure*}
    
    In this section we present the proposed HyperReenact framework for neural face reenactment. An overview of the method is shown in Fig.~\ref{fig:architecture}. In a nutshell, the hypernetwork $\mathcal{H}$, guided by the source appearance ($f_{\mathrm{app}}$) and target facial pose ($f_\mathrm{p}$) related features, learns to predict the offsets $\Delta\theta_{\ell}, \ell={1,\ldots,N}$, where $N$ is the total number of the layers in the generator $\mathcal{G}$. Then, given the initial latent code $\mathbf{w}_s$ and the updated weights $\hat{\theta} = \theta \cdot (1 + \Delta\theta)$, the generator $\mathcal{G}$ is able to generate an image that has the identity of the source and the facial pose of the target face. 
    
    \subsection{GAN Inversion}\label{subsec:gan_inversion}
    
        The generator of StyleGAN2~\cite{karras2020analyzing} takes as input random latent codes $\mathbf{z}\in\mathbb{R}^{512}$ sampled from the standard Gaussian distribution, which are then fed to the mapping network to get the intermediate latent codes $\mathbf{w}\in\mathbb{R}^{512}$ (i.e., codes in the $\mathcal{W}$ space). Most inversion methods (e.g.,~\cite{abdal2019image2stylegan, tov2021designing}) invert to $\mathcal{W}$ space or its extended $\mathcal{W}^+\subseteq\mathbb{R}^{N\times512}$ space, where a different latent code is fed to each of the $N$ layer of the generator. Tov et al.~\cite{tov2021designing} show that, using the $\mathcal{W}^+$ latent space for real image inversion, results in better reconstruction quality. However, it can be challenging to edit the real images using $\mathcal{W}^+$, in particular when altering the head pose, as this can lead to severe visual artifacts (e.g.,~\cite{bounareli2022finding}). In this work, in order to get an initial inverted latent code, we use e4e~\cite{tov2021designing}, which inverts the real images into the $\mathcal{W}^+$ space. We note that we use an off-the-shelf inversion model, trained on the VoxCeleb1~\cite{Nagrani17} dataset and provided by the authors of~\cite{bounareli2022finding}, for the initial inversion step, while during training the inversion encoder $\mathcal{E}_{\mathrm{w}}$ is not updated (see Fig.~\ref{fig:architecture}).

    \subsection{HyperReenact Architecture}\label{subsec:hyperreenact}

        The proposed HyperReenact aims to modify the weights $\theta$ of the generator $\mathcal{G}$ to both lead to better reconstruction (without any further optimization steps) and perform neural face reenactment eliminating any visual artifacts on the generated images (Fig.~\ref{fig:architecture}). Given a pair of source and target images our framework first extracts the corresponding appearance $f_{\mathrm{app}}$ and facial pose $f_{\mathrm{p}}$ features. Specifically, to encode the appearance of a face we use the ArcFace~\cite{deng2019arcface} encoder $E_{\mathrm{app}}$. ArcFace is trained on the face recognition task, as a result its features only capture the identity characteristics of a face. We note that we extract the feature map $f_{\mathrm{app}}$ of shape $512 \times 7 \times 7$ picking the output of the last convolutional layer of ArcFace. Similarly, to encode the facial pose of a face we use the encoder $\mathcal{E}_{\mathrm{p}}$, which is a pretrained 3D shape model (DECA~\cite{feng2020deca}). DECA is trained on 3D facial shape model reconstruction, hence the extracted features only capture the facial pose, without taking into consideration the appearance. We extract the feature map $f_{\mathrm{p}}$ of shape $2048 \times 7 \times 7$, picking the output of the last convolutional layer of $\mathcal{E}_{\mathrm{p}}$. 

        Our goal is to guide the hypernetwork with a feature map that combines the appearance features from the source image and the facial pose features from the target image. Inspired by the Spatially-Adaptive Denormalization (SPADE) module~\cite{park2019semantic}, we propose to \textit{blend} the two feature maps, $f_{\mathrm{app}}$ and $f_{\mathrm{p}}$, using the Reenactment Module (RM). As shown in Fig.~\ref{fig:architecture}, the RM includes a $1 \times 1$ convolution to project $f_{\mathrm{p}}$ into the same channel size as $f_{\mathrm{app}}$, obtaining $f_{\mathrm{p}}^\prime$. Then, for each feature map, we learn two modulation parameters, namely $\gamma$ and $\beta$. The final combined feature map $f_{\mathrm{r}}$ with size $512 \times 7 \times 7$ is calculated as:
        \begin{equation}
            f_{\mathrm{r}} = \gamma_{\mathrm{app}} \odot f_{\mathrm{app}} + \beta_{\mathrm{app}} + \gamma_{\mathrm{p}} \odot f_{\mathrm{p}}' + \beta_{\mathrm{p}},
        \end{equation}
        where $\odot$ denotes the element-wise multiplication.

        Our hypernetwork module follows a similar architecture with the one in HyperStyle~\cite{alaluf2022hyperstyle}. Specifically, it consists of $\mathrm{M} \subset \mathrm{N}$ Reenactment Blocks (RB), where $\mathrm{M}$ is the number of layers we control and $\mathrm{N}$ is the total number of layers of the generator. Each RB takes as input the combined feature map $f_{\mathrm{r}}$ and outputs an offset $\Delta \theta_{\ell}$ of size $C_{\ell}^{out} \times C_{\ell}^{in}\times 1 \times 1$. We then spatially repeat each offset by the kernel dimension of each layer $k_{\ell} \times k_{\ell}$, significantly reducing the number of learnable parameters~\cite{alaluf2022hyperstyle}. Finally, the updated weights for each layer $\ell$ of the generator are computed as $\hat{\theta}_{\ell} = \theta_{\ell} \cdot (1 + \Delta\theta_{\ell})$. We provide a detailed analysis on the architecture of our hypernetwork module in the supplementary material.

    \subsection{Training Process}~\label{ssec:training}

    We train the proposed HyperReenact framework following a curriculum learning (CL) scheme~\cite{bengio2009curriculum}, where we gradually increase the complexity of the training data. Specifically, we first train our network on real image inversion, where the source and target faces are the same. We further train our model on the task of self reenactment, where the source and target faces have the same identity but different facial pose. Finally, we continue training our model on cross-subject reenactment, where the source and target faces have different identity and facial pose. In Section~\ref{ssec:ablation}, we show that the proposed curriculum learning scheme improves our results. We detail each training phase below.

    \noindent \textbf{Phase 1: Real image inversion} On the first training phase, the source ($\mathrm{I}_s$) and the target ($\mathrm{I}_t$) images are the same to the input image ($\mathrm{I}$). Given the appearance and facial pose of the input image $\mathrm{I}$ as well as the initial inverted latent code $\mathbf{w}$, we train our network to refine the missing identity details between the real image and its initial reconstruction using $\mathbf{w}$. Our training objective during the inversion phase consists of the following reconstruction loss terms:
    \begin{equation}\label{eq:loss_inversion}
    \mathcal{L} = \lambda_{pix} \mathcal{L}_{pix} + \lambda_{lpips} \mathcal{L}_{lpips} + \lambda_{id} \mathcal{L}_{id} + \lambda_{g} \mathcal{L}_{g},
    \end{equation}
    where $\mathcal{L}_{pix}$ is the $\ell_1$ pixel-wise loss and $\mathcal{L}_{lpips}$ is the perceptual loss~\cite{johnson2016perceptual} between the real $\mathrm{I}$ and the refined $\hat{\mathrm{I}}$ image. Additionally, to further enhance the identity preservation we calculate the identity loss $\mathcal{L}_{id}$ that computes the cosine similarity of the features extracted using the ArcFace~\cite{deng2019arcface}. Finally, to allow for refinement of the eye gaze direction, we calculate the gaze loss $\mathcal{L}_{g}$, which is the $L2$ distance between the gaze direction of the real image and the reconstructed image $\hat{\mathrm{I}}$, extracted using a gaze estimation method~\cite{zhang2020eth}. To this end, our framework is able to refine the missing identity details from the initial reconstructed images. 

    \noindent \textbf{Phase 2: Self reenactment} During the second training phase, we train our model on self reenactment where the identity of the input source and target images is the same and the facial pose is different. Given the appearance of the source image $\mathrm{I}_s$ and the facial pose of the target image $\mathrm{I}_t$, the hypernetwork is trained to predict the offsets $\Delta\theta$ so that the reenacted image $\mathrm{I}_r$ has the identity of the source image and the facial pose of the target image. The objective during this phase is:
    \begin{equation}\label{eq:loss_self}
        \mathcal{L} = \lambda_{pix} \mathcal{L}_{pix} + \lambda_{lpips} \mathcal{L}_{lpips} + \lambda_{id} \mathcal{L}_{id} + \lambda_{sh} \mathcal{L}_{sh} + \lambda_{g} \mathcal{L}_{g},
    \end{equation}
    where $\mathcal{L}_{pix}$, $\mathcal{L}_{lpips}$, $\mathcal{L}_{id}$, and $\mathcal{L}_{g}$ denote the losses described above, calculated between the target image $\mathrm{I}_t$ and the reenacted image $\mathrm{I}_r$. In order to transfer the facial pose of the target image, we calculate the shape loss $\mathcal{L}_{sh}$, i.e., the $\ell_1$ distance of the 3D facial shapes extracted using~\cite{feng2020deca} from the target and the reenacted image.

    \noindent \textbf{Phase 3: Cross-subject reenactment} To further enhance our results on cross-subject reenactment, we propose to fine-tune our model trained on self reenactment using cross-subject image pairs. As cross-subject training is a challenging task, to ease the training process, we concurrently perform training for the tasks of self and cross-subject reenactment, by letting half of the image pairs in each batch to correspond to each task. Our training objective is the same as the one defined in (\ref{eq:loss_self}), however, for the cross-subject batch samples we only calculate the identity $\mathcal{L}_{id}$ loss between the source and the reenacted images and the shape loss $\mathcal{L}_{sh}$ between the target and the reenacted images.

\section{Experiments}\label{sec:experiments}

    In this section, we provide the implementation details and we present our quantitative and qualitative results and comparisons with state-of-the-art methods. In Section~\ref{ssec:comparisons}, we report results on self and cross-subject reenactment on the VoxCeleb1~\cite{Nagrani17} dataset, and in Section~\ref{ssec:ablation}, we provide ablation studies to investigate the contribution of each design choice to the overall effectiveness of our method. 

    \noindent \textbf{Implementation details} We use the StyleGAN2 model~\cite{karras2020analyzing} and the e4e inversion model~\cite{tov2021designing} trained on the VoxCeleb1 dataset provided by~\cite{bounareli2022finding} and we train our model on the same dataset with $256\times256$ images. We note that the only trainable modules of our method are the the Reenactment Module (RM) and the hypernetwork $\mathcal{H}$ , while the encoders $\mathcal{E}_{\mathrm{w}}$, $\mathcal{E}_{\mathrm{app}}$, and $\mathcal{E}_{\mathrm{p}}$, along with the StyleGAN2 generator $\mathcal{G}$ are kept frozen (Fig.~\ref{fig:architecture}). We also note that we learn offsets for all the layers of the StyleGAN2 generator, except for the ``toRGB'' layers that mainly change the texture and the color of images~\cite{wu2021stylespace} (please see the supplementary material for more details). We first train our model on the task of real image inversion with learning rate $2\cdot10^{-4}$ and we continue on the task of self reenactment with the same learning rate and a batch size of $16$. We finally fine-tune our model on cross-subject reenactment with a constant learning rate of $10^{-4}$. We set $\lambda_{pix} = 10.0$, $\lambda_{lpips} = 5.0$, $\lambda_{id} = 10.0$, $\lambda_{sh} = 0.5$ and $\lambda_{g} = 2.0$. All models are optimized with Adam optimizer~\cite{kingma2014adam} and are implemented in PyTorch~\cite{paszke2017automatic}. 

    \subsection{Comparison with state-of-the-art methods}~\label{ssec:comparisons}

        \begin{table*}[t!]
        \begin{center}
        \begin{tabular}{|l!{\vrule width 1.1pt}c|c|c|c|c|c!{\vrule width 1.1pt}c|c|c!{\vrule width 1.1pt}c|}
            \hline
            \multirow{2}{*}{Method} & \multicolumn{6}{c!{\vrule width 1.2pt}}{Self Reenactment} & \multicolumn{3}{c!{\vrule width 1.2pt}}{Cross-subject Reenactment} & \multirow{2}{*}{\parbox{2cm}{\centering User Pref. ($\%$)}} \\
            \cline{2-10}
             & CSIM$\uparrow$  & LPIPS $\downarrow$ &  FID $\downarrow$ & FVD $\downarrow$ & APD$\downarrow$  & AED$\downarrow$ 
            & CSIM$\uparrow$ & APD$\downarrow$  & AED$\downarrow$ &\\
            \noalign{\hrule height 1.2pt}
            X2Face~\cite{wiles2018x2face} & \underline{0.70} & \textbf{0.21} & \underline{25.6} & 490  &  1.3 & 9.0 & 0.57 & 2.2 & 16.4 & -\\
            FOMM~\cite{siarohin2019first} & 0.64 & 0.27 & 35.3 & 523 &  4.6 & 12.6 & 0.53 & 10.9 & 20.9 & -\\
            Neural~\cite{burkov2020neural} & 0.40 & 0.42  & 127.0 & 617 & 1.2 & 8.8 & 0.34 & 1.8 & 15.3 & - \\
            Fast BL~\cite{zakharov2020fast} & 0.65 & 0.41  & 55.0 & 706 & \underline{1.0} & 7.6 & 0.58 & 1.4 & 14.7 & 5.9 \\  
            PIR~\cite{ren2021pirenderer} & 0.69 & \underline{0.23} & 50.5 & 545 & 1.9 & 9.7 & 0.61 & 2.4 & 15.4 & 1.1  \\
            LSR~\cite{meshry2021learned} & 0.59 & 0.26 & 63.0 & 484 &  \underline{1.0} & 7.5 & 0.50 & 1.5 & 13.1 & 5.0 \\
            FD~\cite{bounareli2022finding} & 0.65 & \underline{0.23} & \textbf{19.0} & \textbf{400} &  \underline{1.0} & 6.5 & 0.49 & 1.7 & 10.2 & 4.1 \\
            LIA~\cite{wang2021latent} & 0.64 & 0.26 & 31.7 & 510 & 4.7 & 11.4 & 0.57 & 2.8 & 15.7 & 5.9 \\
            Dual~\cite{hsu2022dual} & 0.26 & 0.39 & 46.5 &  600 &  3.4 & 12.5 & 0.19 & 3.1 & 16.9 & -\\
            Rome~\cite{khakhulin2022rome} & 0.69 & 0.43 & 39.2 & 800 & 1.5 & \underline{5.6} & \underline{0.63} & \underline{1.2} & \textbf{8.8} & \underline{10.2} \\
            Ours & \textbf{0.71} & \underline{0.23} & 27.1 & \underline{480} & \textbf{0.5} & \textbf{5.1} & \textbf{0.68} & \textbf{0.5} & \underline{9.3} & \textbf{67.8} \\
            \hline
        \end{tabular}
        \end{center}
        \caption{Quantitative results on self and cross-subject reenactment. For CSIM metric, higher is better ($\uparrow$), while for the rest of the metrics lower is better ($\downarrow$). We note that the best and second best results are shown in bold and underline respectively.}\label{table:reenactment_metrics_self_and_cross}
        \end{table*}
        
        \begin{figure*}[t!]
            \centering
            \includegraphics[width=0.8\textwidth]{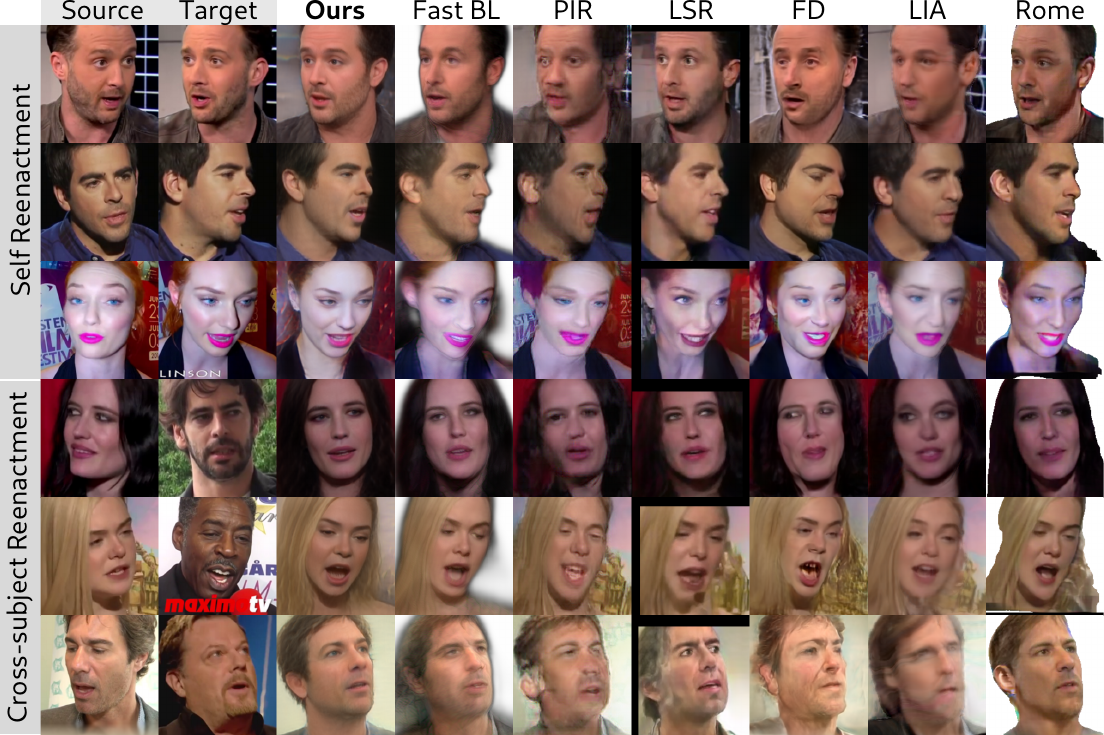}
            \caption{Qualitative results and comparisons on self (first 3 rows) and cross-subject reenactment (last 3 rows) on VoxCeleb1~\cite{Nagrani17}. Our method is able to faithfully preserve the identity characteristics and also effectively transfer the head pose, the facial expression and the gaze direction of the target face without producing significant visual artifacts.}
            \label{fig:comparisons}
        \end{figure*}
        
        We evaluate our method on the test set of the VoxCeleb1~\cite{Nagrani17} dataset and we provide additional quantitative and qualitative results on the test set of the VoxCeleb2~\cite{Chung18b} dataset in the supplementary material. We compare our framework with $10$ state-of-the-art methods that have made their source code and models publicly available, namely, X2Face~\cite{wiles2018x2face}, FOMM~\cite{siarohin2019first}, Neural~\cite{burkov2020neural}, Fast BL~\cite{zakharov2020fast}, PIR~\cite{ren2021pirenderer}, LSR~\cite{meshry2021learned}, FD~\cite{bounareli2022finding}, LIA~\cite{wang2021latent}, Dual~\cite{hsu2022dual}, and Rome~\cite{khakhulin2022rome}. In the supplementary material, we provide additional comparisons with StyleHEAT~\cite{yin2022styleheat} and StyleMask~\cite{bounareli2022StyleMask}. We also note that regarding the methods of Neural~\cite{burkov2020neural}, LSR~\cite{meshry2021learned}, FD~\cite{bounareli2022finding} and Dual~\cite{hsu2022dual}, we perform one-shot fine-tuning on a single source frame from each test video for fair comparisons against the proposed method.

        \noindent \textbf{Quantitative results} We evaluate our method on two tasks, namely, self reenactment and cross-subject reenactment. For self reenactment, we calculate six different evaluation metrics. Specifically, we measure the identity preservation by calculating the cosine similarity (CSIM) of the features extracted using the ArcFace face recognition network~\cite{deng2019arcface}, and the reconstruction quality using the Learned Perceptual Image Path Similarity (LPIPS) metric~\cite{johnson2016perceptual}. Additionally, we calculate the Fr\'echet-Inception Distance (FID)~\cite{heusel2017gans} to measure the quality of the reenacted images and the Fr\'echet-Video Distance (FVD)~\cite{unterthiner2018towards,skorokhodov2022stylegan} to measure the temporal consistency of the generated videos. Finally, to evaluate the facial pose transfer, we calculate the Average Pose Distance (APD) and the Average Expression Distance (AED), similarly to~\cite{ren2021pirenderer}. All metrics for self reenactment are calculated between the target and the reenacted images. When evaluating our method on the task of cross-subject reenactment, we calculate the CSIM metric between the source and the reenacted images, and we also calculate the APD and AED. In Table~\ref{table:reenactment_metrics_self_and_cross}, we report results both on self and on cross-subject reenactment. We note that for self reenactment we report results on the test set of VoxCeleb1, where the first extracted frame from each video is used as the source frame and all other frames are used as the target ones. For cross-subject reenactment, we randomly select $35$ video pairs from the test set of VoxCeleb1. In the task of self reenactment, our method outperforms all other techniques on identity preservation (CSIM), as well as on pose (APD) and facial expression (AED) transfer, while on LPIPS, FID and FVD metrics our method is among the best results. In the more challenging task of cross-subject reenactment, our method outperforms all other techniques on identity preservation and head pose transfer, while being on-par with Rome~\cite{khakhulin2022rome} on expression transfer.

        Moreover, we conduct a user study to further assess the performance of the proposed method in comparison to state-of-the-art works. Specifically, we present 20 \textit{randomly} selected image pairs, 10 of self and 10 of cross-subject reenactment, to 30 users and ask them to select the method that best reenacts the source frame in terms of (i) identity preservation, (ii) facial pose transfer, and (iii) image quality. We note that we opt to include only the methods that exhibit high performance on both quantitative and qualitative results. That is, we exclude X2Face~\cite{wiles2018x2face} and FOMM~\cite{siarohin2019first}, since they lead to several visual artifacts (as shown in the supplementary material). Similarly, we exclude Neural~\cite{burkov2020neural} and Dual~\cite{hsu2022dual} due to their poor quantitative results. As shown in Table~\ref{table:reenactment_metrics_self_and_cross}, our method is by far the most preferable.

        Finally, in order to evaluate the proposed method under the challenging (and far more useful in real-world applications) condition of large variations between the head pose of the source and target frames, we build a small benchmark set for the task of self reenactment, containing pairs of images with large head pose differences. Specifically, from each video of the VoxCeleb1 test set we select 5 image pairs with head pose distance (measured as the average of the absolute differences between the 3 Euler angles) larger than $15^{\circ}$. We report results in Table~\ref{table:large_pose}. Our method outperforms all other techniques on identity preservation and head pose transfer, while ranking second and performing on par with the method of Rome~\cite{khakhulin2022rome} on expression transfer.

        \begin{table}[h]
        \begin{center}
        \begin{tabular}{|l!{\vrule width 1.1pt}c|c|c|c|}
            \hline
            Method & CSIM$\uparrow$ & APD$\downarrow$  & AED$\downarrow$ \\
            \noalign{\hrule height 1.2pt}
            X2Face~\cite{wiles2018x2face} & 0.45 & 3.1 & 12.1 \\
            FOMM~\cite{siarohin2019first} & 0.44 & 3.2 &  12.7 \\
            Neural~\cite{burkov2020neural} & 0.38 & 1.5 &  8.9\\
            Fast BL~\cite{zakharov2020fast} & 0.44 & 1.5 & 9.0 \\
            PIR~\cite{ren2021pirenderer} & 0.40 & 4.0 &  11.9 \\
            LSR~\cite{meshry2021learned} & 0.48 & 1.4 & 8.5 \\
            FD~\cite{bounareli2022finding} & 0.34 & 2.7 & 10.4 \\
            LIA~\cite{wang2021latent} & 0.43 & 3.0 &  9.9 \\
            Dual~\cite{hsu2022dual} & 0.23 & 4.9 & 12.6 \\
            Rome~\cite{khakhulin2022rome} & \underline{0.53} & \underline{1.1} & \textbf{5.8} \\
            Ours & \textbf{0.58} &\textbf{ 0.9} & \underline{6.2}\\
            \hline
        \end{tabular}
        \end{center}
        \caption{Quantitative results on self reenactment using a set of images with large head pose differences between the source and target faces (subset of the VoxCeleb1 test set).}\label{table:large_pose}
        \end{table}

        \noindent \textbf{Qualitative results} In Fig.~\ref{fig:comparisons}, we show qualitative comparisons on self and cross-subject reenactment. We note that for better visualization we report results only with the best performing methods, namely, Fast BL~\cite{zakharov2020fast}, PIR~\cite{ren2021pirenderer}, LSR~\cite{meshry2021learned}, FD~\cite{bounareli2022finding}, LIA~\cite{wang2021latent} and Rome~\cite{khakhulin2022rome}. On the supplementary material we present additional comparisons with all methods. As shown, our method is able to produce mostly artifact free images, successfully preserve the source identity characteristics and faithfully transfer the target facial pose (i.e. head pose orientation, facial expression and gaze direction), even on the challenging task of cross-subject reenactment and on extreme head pose differences.  

    \subsection{Ablation studies}~\label{ssec:ablation}
        
\begin{figure}[t!]
            \centering
            \includegraphics[width=0.8\linewidth]{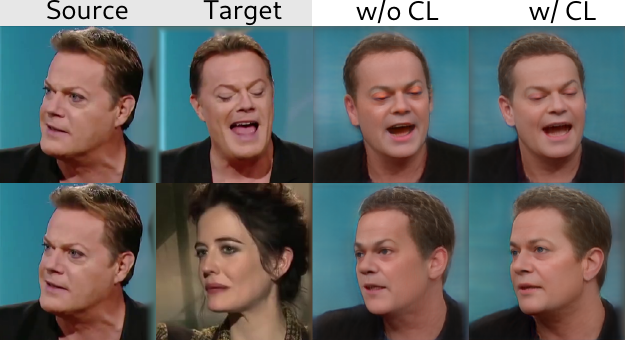}
            \caption{Qualitative results on self and cross-subject reenactment with and without using the proposed CL scheme.}
            \label{fig:ablation_curriculum}
        \end{figure}

        In this section, we report the results of the ablation studies we performed to assess the contribution of: (a) the proposed curriculum learning (CL) scheme, (b) fine-tuning on the task of cross-subject reenactment, and (c) the gaze loss. Regarding (a), i.e. the contribution of the proposed curriculum learning scheme (Section~\ref{ssec:training}), we provide results both on self and on cross-subject reenactment with our final model trained using the curriculum learning (CL) scheme and with a model trained directly on the task of self and cross-subject reenactment. As shown in Table.~\ref{table:ablation_curriculum}, our model trained with CL achieves better results on head pose (APD), expression transfer (AED), and identity preservation on self reenactment. This is also shown in Fig.~\ref{fig:ablation_curriculum}, where using CL leads to results less visual artifacts on the reenacted images.
        
        \begin{table}[h]
        \begin{center}
        \resizebox{\columnwidth}{!}{%
        \begin{tabular}{|p{1.1cm}!{\vrule width 1.2pt}c|c|c!{\vrule width 1.2pt}c|c|c|}
            \hline
             \multirow{2}{*}{Method} & \multicolumn{3}{c!{\vrule width 1.2pt}}{Self Reenactment} & \multicolumn{3}{c|}{Cross Reenactment} \\
             \cline{2-7} 
             & CSIM & APD & AED & CSIM & APD  & AED \\
            \noalign{\hrule height 1.2pt}
            \textit{w/o} CL  & 0.69 & 0.7 & 6.7 & \textbf{0.72} & 0.6 & 11.6 \\
            \textit{w/} CL & \textbf{0.71} & \textbf{0.5} & \textbf{5.1} & 0.68 & \textbf{0.5} & \textbf{9.3} \\
            \hline
        \end{tabular}
        }
        \end{center}
        \caption{Quantitative results on self and cross-subject reenactment with and without using curriculum learning (CL).}\label{table:ablation_curriculum}
        \end{table}
        
\begin{figure}[t!]
            \centering
            \includegraphics[width=0.8\linewidth]{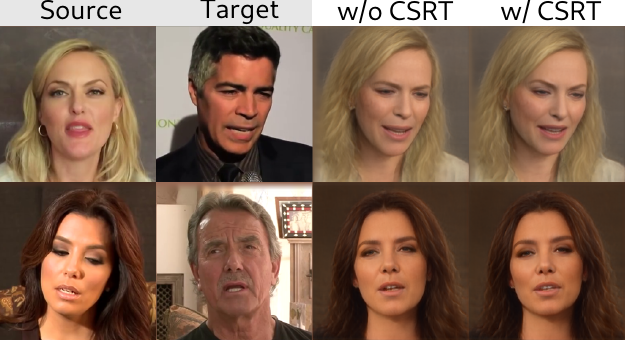}
            \caption{Qualitative results on cross-subject reenactment with and without training with cross-subject data (CSRT).}
            \label{fig:ablation_cross}
        \end{figure}

        For (b), we compare our method with and without training with cross-subject data. As shown in Table~\ref{table:ablation_cross}, when fine-tuning our method with cross-subject data, our quantitative results with respect to identity preservation (CSIM) are improved especially on cross-subject reenactment. The effect of cross-subject training on the identity preservation is also shown in Fig.~\ref{fig:ablation_cross}, where we eliminate the identity leakage from the target images into the reenacted ones.
        
        \begin{table}[h]
        \begin{center}
        \begin{tabular}{|c!{\vrule width 1.2pt}c|c|}
            \hline
            \multirow{2}{*}{Method} & \multicolumn{2}{c|}{CSIM$\uparrow$} \\
             \cline{2-3} 
             & SR & CR  \\
            \noalign{\hrule height 1.2pt}
            \textit{w/o} CSRT & 0.69 & 0.53  \\
            \textit{w/} CSRT & \textbf{0.71} & \textbf{ 0.68} \\
            \hline
        \end{tabular}
        \end{center}
        \caption{Quantitative results on self (SR) and cross-subject (CR) reenactment with and without performing cross-subject reenactment training (CSRT).}\label{table:ablation_cross}
        \end{table}
        
\begin{figure}[t!]
            \centering
            \includegraphics[width=0.8\linewidth]{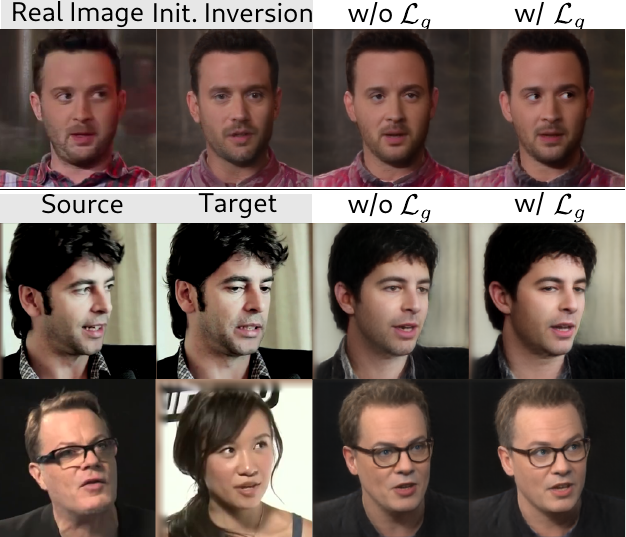}
            \caption{Illustration of the impact of the gaze loss $\mathcal{L}_{g}$ on the generated images, on the tasks of real image inversion and self/cross-subject reenactment.}
            \label{fig:gaze_loss}
        \end{figure}

        For (c), we perform experiments with and without using the gaze loss $\mathcal{L}_{g}$ and we evaluate its contribution both quantitatively (Table~\ref{table:ablation_gaze}) and qualitatively (Fig.~\ref{fig:gaze_loss}). Specifically, in Table~\ref{table:ablation_gaze} we calculate the Gaze Error, i.e., the $\ell_2$ distance between the gaze direction of the real and the generated images on three different tasks, namely, real image inversion (I), self reenactment (SR), and cross-subject reenactment (CR). Moreover, in Fig.~\ref{fig:gaze_loss} we present examples of real image inversion (first row) and self and cross-subject reenactment (last two rows), with and without the gaze loss, where we observe that using the gaze loss improves the faithful reconstruction of the gaze direction of the target images.
        
        \begin{table}[h]
        \begin{center}
        \begin{tabular}{|c!{\vrule width 1.1pt}c|c|c|}
            \hline
            \multirow{2}{*}{Method} & \multicolumn{3}{c|}{Gaze Error $\downarrow$} \\
             \cline{2-4}    
             & I & SR & CR  \\
            \noalign{\hrule height 1.2pt}
            \textit{w/o} $\mathcal{L}_{g}$ & 0.34 & 0.35 & 0.40  \\
            \textit{w/} $\mathcal{L}_{g}$ & \textbf{0.24} & \textbf{0.25} & \textbf{0.31}\\
            \hline
        \end{tabular}
        \end{center}
        \caption{Ablation on the impact of gaze loss $\mathcal{L}_{g}$. The gaze error is calculated on real image inversion (I), self reenactment (SR) and cross-subject reenactment (CR) tasks.}\label{table:ablation_gaze}
        \end{table}
        
\section{Conclusions}\label{sec:conclusions}

    In this paper, we present HyperReenact, a method for neural face reenactment that jointly learns to refine and re-target the facial images using a pretrained StyleGAN2 model and a hypernetwork. We leverage the effectiveness of hypernetworks on the real image inversion task and extend their use for real image manipulation. Our method learns to fuse the disentangled representations of source identity and target facial pose, to effectively modify the weights of the generator, performing both identity refinement and facial pose re-targeting. We show that our approach can successfully reenact a source face, preserving the identity and transferring the target facial pose. We also demonstrate that our method can produce artifact-free images even on challenging cases of extreme head pose movements.

{\setlength{\parindent}{0cm}
\textbf{Acknowledgments:} This work was supported by the EU H2020 AI4Media No. 951911 project.
}

\clearpage
\appendix
\section{Supplementary Material}\label{supp}

In this supplementary material, we first provide a detailed analysis of the network architecture of the proposed framework (HyperReenact) in Section~\ref{sec:architecture}. In Section~\ref{sec:hyperstyle}, we show results of two state-of-the-art inversion methods, namely, HyperStyle~\cite{alaluf2022hyperstyle} and HyperInverter ~\cite{dinh2022hyperinverter}, on real image editing and compare our method against HyperStyle~\cite{alaluf2022hyperstyle}. In Section~\ref{sec:limitation}, we discuss the limitations of our method and in Section~\ref{sec:styleheat} we present comparisons against two methods, namely, StyleHEAT~\cite{yin2022styleheat} and StyleMask~\cite{bounareli2022StyleMask}, while in Section~\ref{sec:results} we provide additional results of our method on the VoxCeleb1~\cite{Nagrani17} and the VoxCeleb2~\cite{Chung18b} datasets. Finally, along with this report, please find attached an external video file that includes 10 randomly selected identities for self reenactment and 10 randomly selected pairs for cross-subject reenactment (for both VoxCeleb1~\cite{Nagrani17} and VoxCeleb2~\cite{Chung18b} datasets).

    \subsection{Network architecture}~\label{sec:architecture}
    
\begin{figure}[t]
            \centering
            \includegraphics[width=1.0\linewidth]{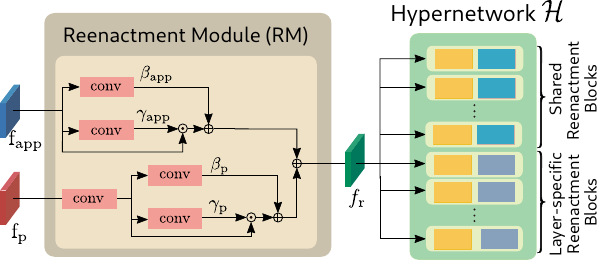}
            \caption{Illustration of the two learnable components of our architecture, namely, the Reenactment Module (RM) and the hypernetwork $\mathcal{H}$. The RM module fuses appearance features from the source face and pose features from the target face and outputs the fused feature map $f_{\mathrm{r}}$, that drives the hypernetwork $\mathcal{H}$. The hypernetwork $\mathcal{H}$ consists of multiple Reenactment Blocks with each one of them corresponding to a particular layer of the generator.}
            \label{fig:reenactment_module}
        \end{figure}

        In this section we provide details of the various components of the proposed framework (HyperReenact). An overview of HyperReenact is shown in Fig.~\ref{fig:architecture} in the main paper. More specifically, we propose to blend the appearance feature map $f_{\mathrm{app}}$ of size $512 \times 7 \times 7$ and the pose feature map $f_{\mathrm{p}}$ of size $2048 \times 7 \times 7$ using the Reenactment Module (RM), which is illustrated in Fig.~\ref{fig:reenactment_module}. As shown, we first project the $f_{\mathrm{p}}$ using a convolutional layer (kernel=(1,1), stride=1, pad=0) into the same channel size of $f_{\mathrm{app}}$. Then for each feature map we learn the two modulation parameters $\gamma$ and $\beta$ using convolutions with a kernel size of 1, stride set to 1 and padding set to 0. The output feature map $f_{\mathrm{r}}$ of size $512 \times 7 \times 7$, computed using Eq.~1 from the main paper, is then fed into the different reenactment blocks of the hypernetwork. 
    
        Our hypernetwork $\mathcal{H}$ consists of $\mathrm{M}<\mathrm{N}$ Reenactment Blocks (RB), where $\mathrm{M}$ is the number of generator layers that we control and $\mathrm{N}$ is the total number of layers in the generator. Each of those blocks takes as input the feature map $f_{\mathrm{r}}$ and outputs the corresponding offset $\Delta \theta_{\ell}$ for the weights of each layer of the generator. In Table~\ref{table:stylegan2_arch} we report the structure of StyleGAN2 generator trained on $256 \times 256$ image resolution. From the blocks shown in Table~\ref{table:stylegan2_arch} we only modify the convolutional layers (Conv), as the ToRGB layers mainly affect the colors of the generated images~\cite{wu2021stylespace}. Hence, we propose to modify $\mathrm{M} = 13$ layers. 
    
\begin{figure*}[t]
            \centering
            \includegraphics[width=1.0\linewidth]{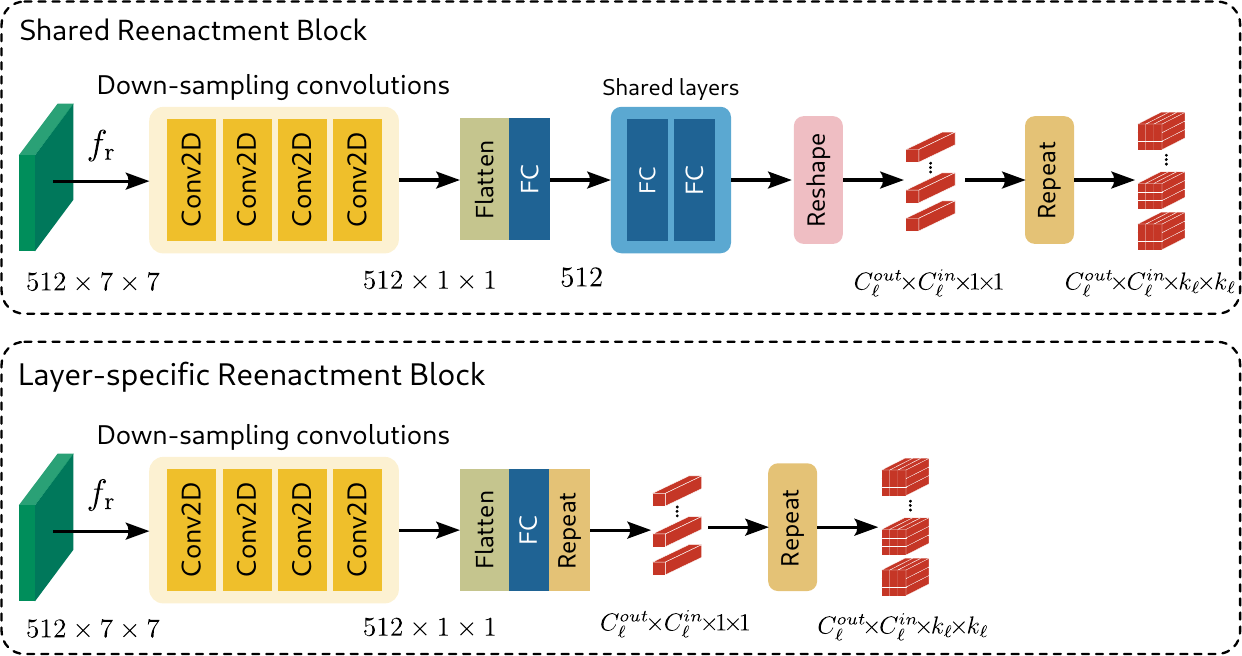}
            \caption{Detailed architecture of the two types of Reenactment Blocks, namely the Shared Reenactment Block (top row) and the Layer-specific Reenactment Block (bottom row).}
            \label{fig:architecture_details}
        \end{figure*}

        In order to reduce the number of trainable parameters, similarly to HyperStyle~\cite{alaluf2022hyperstyle}, we propose to use two types of Reenactment Blocks, namely, the Shared Reenactment Blocks and the Layer-specific Reenactment Blocks, as shown in Fig.~\ref{fig:reenactment_module}. We present an overview of the architecture of both blocks in Fig.~\ref{fig:architecture_details}. In both types of blocks, the input feature map $f_{\mathrm{r}}$ is first fed into a series of convolutional layers that process and down-sample the input into the shape of $512 \times 1 \times 1$. Regarding the Shared Reenactment Blocks, as shown on the top row of Fig.~\ref{fig:architecture_details}, the down-sampled feature map is then flattened and fed into a fully-connected layer. Afterwards, the two shared fully-connected layers are used to calculate the output feature map of shape $C_{\ell}^{out} \times C_{\ell}^{in} \times 1 \times 1$, which is repeated spatially so as to match the shape of the convolutional kernels ($C_{\ell}^{out} \times C_{\ell}^{in} \times k_{\ell} \times k_{\ell}$). Regarding the Layer-specific Reenactment Blocks, as shown on the bottom row of Fig.~\ref{fig:architecture_details}, after the series of down-sampling convolutional layers, the computed feature map has a shape of $512 \times 1 \times 1$. This feature map, upon being flattened, is fed into a final fully-connected layer which outputs a feature map of shape $C_{\ell}^{out} \times C_{\ell}^{in} \times 1 \times 1$, which is then spatially repeated to have a shape of $C_{\ell}^{out} \times C_{\ell}^{in} \times k_{\ell} \times k_{\ell}$. 
    
        We provide in more detail the structure of the two Reenactment Blocks in Tables~\ref{table:shared_info} and~\ref{table:spec_info}. Finally, in Table~\ref{table:modified_layers}, we report the StyleGAN2 layers that we propose to modify as well as the type of the Reenactment Block that we use for each layer. As shown in Table~\ref{table:modified_layers}, we use the Shared Reenactment Blocks for the first seven layers of the generator. As a result, the parameters of two fully-connected layers across all the Shared Reenactment Blocks, are common. For the last six layers of the generator, we use the Layer-specific Reenactment Blocks. We note that the use of the shared layers reduces our trainable parameters from $1.2\mathrm{B}$ to $300\mathrm{M}$.
    
        \begin{table*}[t]
        \setlength{\tabcolsep}{5pt}
        \begin{center}
        \begin{tabular}{P{2cm} | P{2cm} |  P{2cm} | P{5cm}} 
             \toprule
            Layer $\ell$ Index & Layer $\ell$ Name & Resolution & Kernel Dim. $C_{\ell}^{out} \times C_{\ell}^{in} \times k_{\ell} \times k_{\ell}$ \\  
             \midrule
             0 & Conv1 & $4 \times 4$ & $512 \times 512 \times 3 \times 3$ \\
             1 & ToRGB1 & $4 \times 4$ &  $3 \times 512 \times 1 \times 1$ \\
             \hline
             2 & Conv2 & $8 \times 8$ &  $512 \times 512 \times 3 \times 3$ \\
             3 & Conv3 & $8 \times 8$ & $512 \times 512 \times 3 \times 3$ \\
             4 & ToRGB2 & $8 \times 8$ &  $3 \times 512 \times 1 \times 1$ \\
             \hline
             5 & Conv4 & $16 \times 16$ & $512 \times 512 \times 3 \times 3$ \\
             6 & Conv5 & $16 \times 16$ & $512 \times 512 \times 3 \times 3$ \\
             7 & ToRGB3 & $16 \times 16$ & $3 \times 512 \times 1 \times 1$ \\
             \hline
             8 & Conv6 & $32 \times 32$ & $512 \times 512 \times 3 \times 3$ \\
             9 & Conv7 & $32 \times 32$ & $512 \times 512 \times 3 \times 3$ \\
             10 & ToRGB4 & $32 \times 32$ & $3 \times 512 \times 1 \times 1$ \\
             \hline
             11 & Conv8 & $64 \times 64$ & $256 \times 512 \times 3 \times 3$ \\
             12 & Conv9 & $64 \times 64$ & $256 \times 256 \times 3 \times 3$ \\
             13 & ToRGB5 & $64 \times 64$ & $3 \times 256 \times 1 \times 1$ \\
             \hline
             14 & Conv10 & $128 \times 128$ & $128 \times 256 \times 3 \times 3$ \\
             15 & Conv11 & $128 \times 128$ & $128 \times 128 \times 3 \times 3$ \\
             16 & ToRGB6 & $128 \times 128$ & $3 \times 128 \times 1 \times 1$ \\
             \hline
             17 & Conv12 & $256 \times 256$ & $64 \times 128 \times 3 \times 3$ \\
             18 & Conv13 & $256 \times 256$ & $64 \times 64 \times 3 \times 3$ \\
             19 & ToRGB7 & $256 \times 256$ & $3 \times 64 \times 1 \times 1$ \\
             \midrule
        \end{tabular}
        \caption{Structure of StyleGAN2 generator trained on $256\times 256$ image resolution.}\label{table:stylegan2_arch}
        \end{center}
        \end{table*}

        \begin{table*}
        \centering
        \begin{tabular}{c|c|c}
        \toprule
        Layer & Input shape & Output shape \\
        \hline
        \begin{tabular}[c]{@{}c@{}}Conv2D\\kernel=(3,3), stride=1, pad=1\end{tabular} & $B \times 512 \times 7 \times 7$ &  $B \times 128 \times 7 \times 7$ \\
        \hline
         \begin{tabular}[c]{@{}c@{}}LeakyReLU activation\\slope=0.01\end{tabular} & $B \times 128 \times 7 \times 7$ & $B \times 128 \times 7 \times 7$  \\
         \hline
         \begin{tabular}[c]{@{}c@{}}Conv2D\\kernel=(3,3), stride=1, pad=0\end{tabular} & $B \times 128 \times 7 \times 7$ & $B \times 128 \times 5 \times 5$ \\
         \hline
         \begin{tabular}[c]{@{}c@{}}LeakyReLU activation\\slope=0.01\end{tabular} & $B \times 128 \times 5 \times 5$ & $B \times 128 \times 5 \times 5$  \\
         \hline
         \begin{tabular}[c]{@{}c@{}}Conv2D\\kernel=(3,3), stride=1, pad=0\end{tabular} & $B \times 128 \times 5 \times 5$ & $B \times 128 \times 3 \times 3$ \\
         \hline
         \begin{tabular}[c]{@{}c@{}}LeakyReLU activation\\slope=0.01\end{tabular} & $B \times 128 \times 3 \times 3$ & $B \times 128 \times 3 \times 3$  \\
         \hline
         \begin{tabular}[c]{@{}c@{}}Conv2D\\kernel=(3,3), stride=1, pad=0\end{tabular} & $B \times 128 \times 3 \times 3$ & $B \times 512 \times 1 \times 1$ \\
         \hline
         \begin{tabular}[c]{@{}c@{}}LeakyReLU activation\\slope=0.01\end{tabular} & $B \times 512 \times 1 \times 1$ & $B \times 512 \times 1 \times 1$  \\
         \hline
         Flatten & $B \times 512 \times 1 \times 1$ & $B \times 512 $  \\
         \hline
         FC & $B \times 512$ & $B \times 512 $  \\
         \hline
         Shared FC & $B \times 512$ & $B \times (512\times512) $  \\
         \hline
         Shared FC & $B \times (512\times512)$ & $B \times (512\times512)$\\ 
         \hline
         Reshape & $B \times (512\times512)$ & $B \times 512 \times 512 \times 1 \times 1 $\\
         \hline
         Repeat & $B \times 512 \times 512 \times 1 \times 1 $ & $B \times 512 \times 512 \times 3 \times 3 $\\
         \midrule
        \end{tabular}\label{table:layer_sp_blocks}
        \caption{Architecture of the Shared Reenactment Blocks ($B$ denotes the batch size).}\label{table:shared_info}
        \end{table*}
        
        \begin{table*}
        \centering
        \begin{tabular}{c|c|c}
        \toprule
        Layer & Input shape & Output shape \\
        \hline
        \begin{tabular}[c]{@{}c@{}}Conv2D\\kernel=(3,3), stride=1, pad=1\end{tabular} & $B \times 512 \times 7 \times 7$ &  $B \times 256 \times 7 \times 7$ \\
        \hline
        \begin{tabular}[c]{@{}c@{}}LeakyReLU activation\\slope=0.01\end{tabular} & $B \times 256 \times 7 \times 7$ & $B \times 256 \times 7 \times 7$  \\
        \hline
        \begin{tabular}[c]{@{}c@{}}Conv2D\\kernel=(3,3), stride=1, pad=0\end{tabular} & $B \times 256 \times 7 \times 7$ & $B \times 256 \times 5 \times 5$ \\
        \hline
        \begin{tabular}[c]{@{}c@{}}LeakyReLU activation\\slope=0.01\end{tabular} & $B \times 256 \times 5 \times 5$ & $B \times 256 \times 5 \times 5$  \\
        \hline
        \begin{tabular}[c]{@{}c@{}}Conv2D\\kernel=(3,3), stride=1, pad=0\end{tabular}& $B \times 256 \times 5 \times 5$ & $B \times 256 \times 3 \times 3$ \\
        \hline
        \begin{tabular}[c]{@{}c@{}}LeakyReLU activation\\slope=0.01\end{tabular} & $B \times 256 \times 3 \times 3$ & $B \times 256 \times 3 \times 3$  \\
        \hline
        \begin{tabular}[c]{@{}c@{}}Conv2D\\kernel=(3,3), stride=1, pad=0\end{tabular} & $B \times 256 \times 3 \times 3$ & $B \times 512 \times 1 \times 1$ \\
        \hline
        \begin{tabular}[c]{@{}c@{}}LeakyReLU activation\\slope=0.01\end{tabular} & $B \times 512 \times 1 \times 1$ & $B \times 512 \times 1 \times 1$  \\
        \hline
        Flatten & $B \times 512 \times 1 \times 1$ & $B \times 512 $  \\
        \hline
        FC & $B \times 512$ & $B \times (C_{\ell}^{out} \times C_{\ell}^{in}) $ \\
        \hline
        Reshape & $B \times (C_{\ell}^{out} \times C_{\ell}^{in}) $ & $B \times C_{\ell}^{out} \times C_{\ell}^{in} \times 1 \times 1 $\\
         \hline
        Repeat & $B \times C_{\ell}^{out} \times C_{\ell}^{in} \times 1 \times 1$ & $B \times C_{\ell}^{out} \times C_{\ell}^{in} \times 3 \times 3$ \\
         \midrule
        \end{tabular}\label{table:layer_sp_blocks}
        \caption{Architecture of the Layer-specific Reenactment Blocks. The input on the block has a size of $B \times 512 \times 7 \times 7$, where $B$ is the batch size, while $C_{\ell}^{in}$ and $C_{\ell}^{out}$ are the input and output channels, respectively.}\label{table:spec_info}
        \end{table*}
        
        \begin{table}[t]
        \setlength{\tabcolsep}{5pt}
        \begin{center}
        \begin{tabular}{P{2cm} | P{2cm} |  P{2cm} } 
             \toprule
            Layer Index $\ell$ & $\ell$-th Layer Name & RB Type \\  
             \midrule
             0 & Conv1 & Shared \\
             \hline
             2 & Conv2 & Shared \\
             3 & Conv3 & Shared \\
             \hline
             5 & Conv4 & Shared \\
             6 & Conv5 & Shared\\
             \hline
             8 & Conv6 & Shared\\
             9 & Conv7 & Shared \\
             \hline
             11 & Conv8 & Layer-specific \\
             12 & Conv9 & Layer-specific \\
             \hline
             14 & Conv10 & Layer-specific  \\
             15 & Conv11 & Layer-specific\\
             \hline
             17 & Conv12 & Layer-specific\\
             18 & Conv13 & Layer-specific \\
             \midrule
        \end{tabular}
        \end{center}
        \caption{Convolutional layers of the StyleGAN2 generator that we propose to modify, altering their weights using the offsets computed by the hypernetwork.}\label{table:modified_layers}
        \end{table}

    \subsection{Comparisons with HyperStyle}~\label{sec:hyperstyle}
        
        As discussed in the main paper, the proposed framework draws inspiration from two state-of-the-art methods (for the task of real image inversion), namely, HyperStyle~\cite{alaluf2022hyperstyle} and HyperInverter~\cite{dinh2022hyperinverter}. Specifically, similarly to~\cite{alaluf2022hyperstyle,dinh2022hyperinverter}, we also incorporate a hypernetwork~\cite{hypernets} in order to learn how to effectively modify the weights of a pretrained StyleGAN2~\cite{karras2020analyzing} generator. However, we note that~\cite{alaluf2022hyperstyle,dinh2022hyperinverter} aim at the problem of real image inversion, not neural face reenactment. 
        
\begin{figure}[t]
            \begin{subfigure}{.45\textwidth}
                \centering
                \includegraphics[width=.8\linewidth]{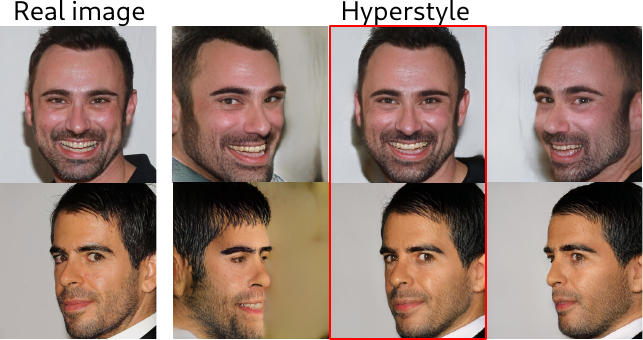}
                \caption{HyperStyle~\cite{alaluf2022hyperstyle}.}
                \label{fig:sub_hyperstyle}
            \end{subfigure}
            \begin{subfigure}{.45\textwidth}
                \centering
                \includegraphics[width=.8\linewidth]{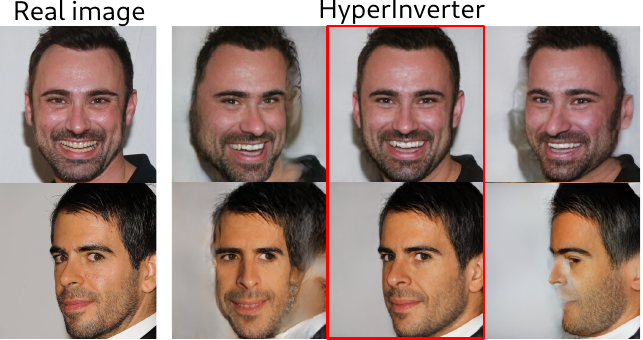}
                \caption{HyperInverter~\cite{dinh2022hyperinverter}.}
                \label{fig:sub_hyperinverter}
            \end{subfigure}
            \caption{Real image inversion and editing results on CelebA dataset~\cite{karras2019style} using HyperStyle~\cite{alaluf2022hyperstyle} and HyperInverter~\cite{dinh2022hyperinverter}. Inside the red boxes are the inverted images, while on the right and left we show results of head pose editing using the method of InterFaceGAN~\cite{shen2020interfacegan}.}
            \label{fig:hyperstyle_hyperinverter}
        \end{figure}

        Both HyperStyle~\cite{alaluf2022hyperstyle} and HyperInverter~\cite{dinh2022hyperinverter} produce high-quality results on real image inversion, however their quality degenerates significantly when manipulating the inverted images, especially on head pose editing. In Fig.~\ref{fig:hyperstyle_hyperinverter}, we show results of HyperStyle~\cite{alaluf2022hyperstyle} (Fig.~\ref{fig:sub_hyperstyle}) and HyperInverter~\cite{dinh2022hyperinverter} (Fig.~\ref{fig:sub_hyperinverter}) on head pose editing using the the CelebA dataset~\cite{karras2019style}. It is worth noting that, while both methods excel on real image inversion (the inverted images are inside the red boxes), they produce several visual artifacts on the edited images, which renders them unsuitable for the task of real image reenactment. We note that we obtain the edited images using the InterFaceGAN method~\cite{shen2020interfacegan} to shift the latent codes.
        
\begin{figure}[t]
            \centering
            \includegraphics[width=0.9\linewidth]{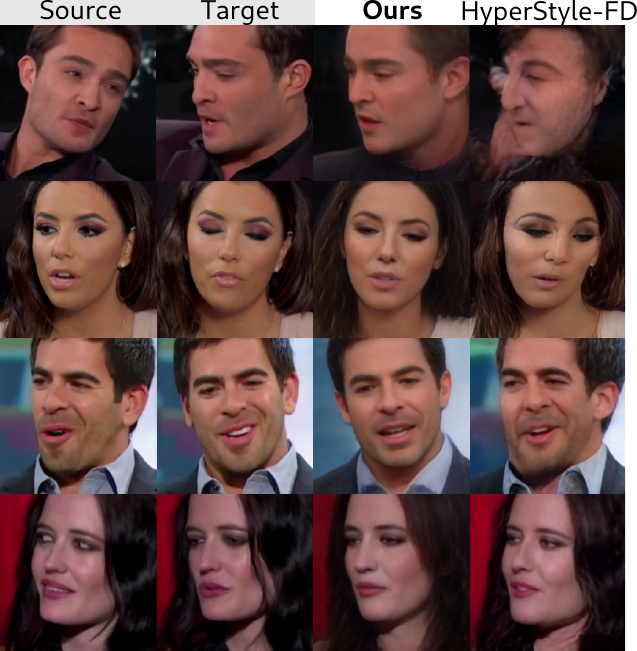}
            \caption{Qualitative comparisons against HyperStyle-FD.}
            \label{fig:hyperstyle}
        \end{figure}

        To further compare our method with HyperStyle~\cite{alaluf2022hyperstyle}, instead of simply editing the head pose as shown above, we perform face reenactment using the learned directions from the work of FD~\cite{bounareli2022finding}. We note that FD learns the directions in the $W^+$ latent space of a StyleGAN2 model trained on the VoxCeleb1~\cite{Nagrani17} dataset that are responsible for controlling the facial pose. In order to test HyperStyle along with FD, we train HyperStyle on the VoxCeleb1 dataset. We refer to this pipeline as HyperStyle-FD. In Fig.~\ref{fig:hyperstyle} we show comparisons of our method against HyperStyle-FD. We note that the reenacted images using HyperStyle-FD not only present visual artifacts, but also look unnatural, especially when the source and target images have large head pose differences.
    
    \subsection{Limitations}~\label{sec:limitation}
    
\begin{figure}[t]
            \centering
            \includegraphics[width=0.7\linewidth]{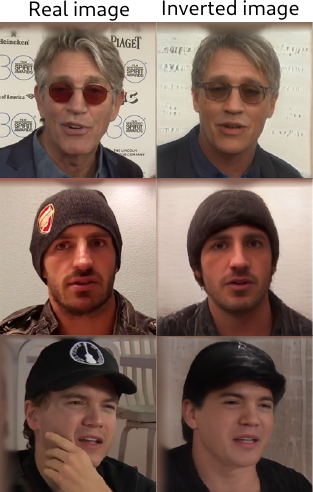}
            \caption{We observe that accessory items such as hats or glasses are not fully reconstructed.}
            \label{fig:limitations}
        \end{figure}

        As shown in the main paper and in the additional experimental results provided in this supplementary material, the proposed HyperReenact, in contrast to several state-of-the-art works, achieves to effectively reenact a source face given a target facial pose, preserving the source identity characteristics and producing artifact-free images, especially in the cases where the target and the source faces differ largely in head pose. Nevertheless, we observe that in cases where the source facial images depict accessories such hats or eyeglasses, the proposed method fails to reconstruct them in detail. For instance, as shown in Fig.~\ref{fig:limitations}, our method cannot fully reconstruct the style of the glasses in the example of the first row. Similarly, regarding the examples of the second and third row of Fig.~\ref{fig:limitations}, our method is not able to reconstruct every detail on the hats. We attribute this to the fact that such items are underrepresented on the VoxCeleb1 dataset and, as a result, our method is not able to learn how to reconstruct them. Additionally, we do not refine any details on the background of the generated images. 
    
    \subsection{Comparisons with StyleHEAT~\cite{yin2022styleheat} and StyleMask~\cite{bounareli2022StyleMask}}~\label{sec:styleheat}
    
\begin{figure}[t]
            \centering
            \includegraphics[width=1.0\linewidth]{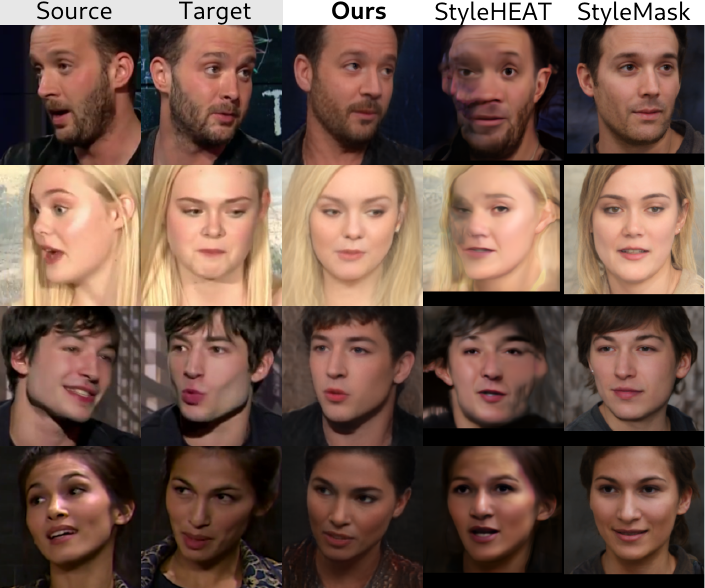}
            \caption{Qualitative comparisons with StyleHEAT~\cite{yin2022styleheat} and StyleMask~\cite{bounareli2022StyleMask} on VoxCeleb dataset~\cite{Nagrani17}.}
            \label{fig:styleheat_vox}
        \end{figure}

\begin{figure}[t]
            \centering
            \includegraphics[width=1.0\linewidth]{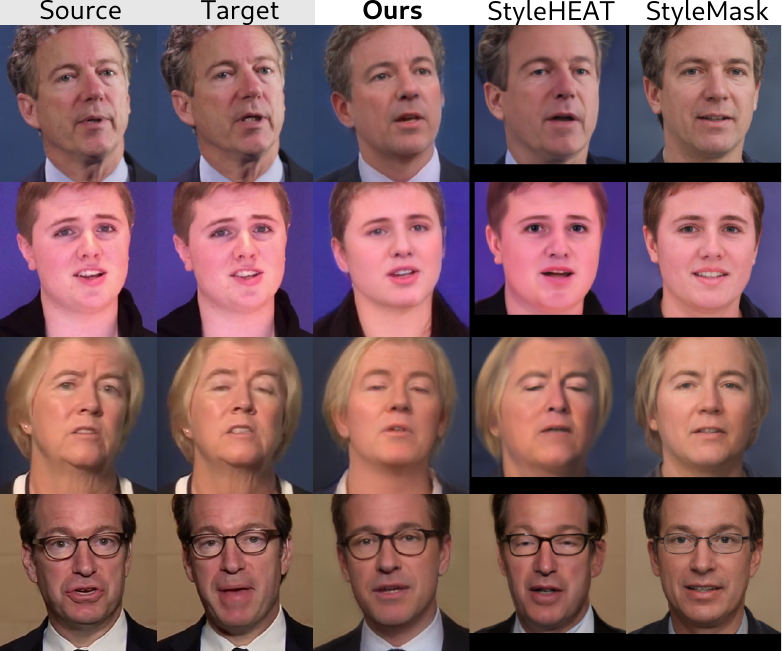}
            \caption{Qualitative comparisons with StyleHEAT~\cite{yin2022styleheat} and StyleMask~\cite{bounareli2022StyleMask} on HDTF dataset~\cite{zhang2021flow}.}
            \label{fig:styleheat}
        \end{figure}

        As discussed in Section~\ref{sec:experiments}, StyleHEAT~\cite{yin2022styleheat} is trained on the HDTF dataset~\cite{zhang2021flow}, that consists of facial images exhibiting only small roll angle variations and showing mostly frontal views. Moreover, StyleMask~\cite{bounareli2022StyleMask} is a face reenactment method based on a pretrained StyleGAN2 model trained on the FFHQ dataset, which learns to disentangle the identity characteristics from the facial pose using the disentangled properties of the style space $\mathcal{S}$ of StyleGAN2. Both StyleHEAT and StyleMask require the input images to be aligned, similarly to the FFHQ dataset~\cite{karras2019style}. In Fig.~\ref{fig:styleheat_vox} and in Table~\ref{table:styleheat_vox}, we present qualitative and quantitative comparisons with StyleHEAT and StyleMask on the VoxCeleb1 dataset~\cite{Nagrani17}. Clearly, StyleHEAT performs poorly by generating many visual artifacts when the source and target images have large pose variations, while StyleMask is not able to faithfully reconstruct the source identity characteristics. For a fair comparison, we additionally compare on the HDTF dataset~\cite{zhang2021flow} (where StyleHEAT has been trained on). In Table~\ref{table:styleheat}, we provide quantitative results on the test videos provided by the authors of StyleHEAT~\cite{yin2022styleheat}. Finally, Fig.~\ref{fig:styleheat} illustrates qualitative comparisons with both StyleHEAT and StyleMask. As shown, our method evidently outperforms both StyleHEAT and StyleMask, on identity preservation and on facial pose transfer.
        
        \begin{table}[t]
        \begin{center}
        \begin{tabular}{|l!{\vrule width 1.1pt}c|c|c|}
            \hline
            Method & CSIM$\uparrow$  & APD$\downarrow$  & AED$\downarrow$ \\
            \noalign{\hrule height 1.2pt}
            StyleHEAT~\cite{yin2022styleheat} & 0.45 & 8.6 & 12.9 \\
            StyleMask~\cite{bounareli2022StyleMask} & 0.47 & 5.3 & 13.2 \\
            Ours & \textbf{0.58} & \textbf{0.9} & \textbf{6.2}  \\
            \hline
        \end{tabular}
        \end{center}
        \caption{Quantitative comparisons with StyleHEAT~\cite{yin2022styleheat} and StyleMask~\cite{bounareli2022StyleMask} on the small benchmark with large head pose differences between the source and target faces.}\label{table:styleheat_vox}
        \end{table}
    
        \begin{table}[t]
        \begin{center}
        \begin{tabular}{|l!{\vrule width 1.1pt}c|c|c|}
            \hline
            Method & CSIM$\uparrow$  & APD$\downarrow$  & AED$\downarrow$ \\
            \noalign{\hrule height 1.2pt}
            StyleHEAT~\cite{yin2022styleheat} & 0.72 & 1.1 & 7.5 \\
            StyleMask~\cite{bounareli2022StyleMask} & 0.66 & 1.6 & 8.8 \\
            Ours & \textbf{0.75} & \textbf{0.38} & \textbf{4.1}  \\
            \hline
        \end{tabular}
        \end{center}
        \caption{Quantitative comparisons on self-reenactment with StyleHEAT~\cite{yin2022styleheat} and StyleMask~\cite{bounareli2022StyleMask} on HDTF dataset~\cite{zhang2021flow}}\label{table:styleheat}
        \end{table}
        
    \subsection{Benchmark with extreme head pose differences}\label{sec:extreme_pose}
    
\begin{figure*}[t]
            \centering
            \includegraphics[width=0.9\linewidth]{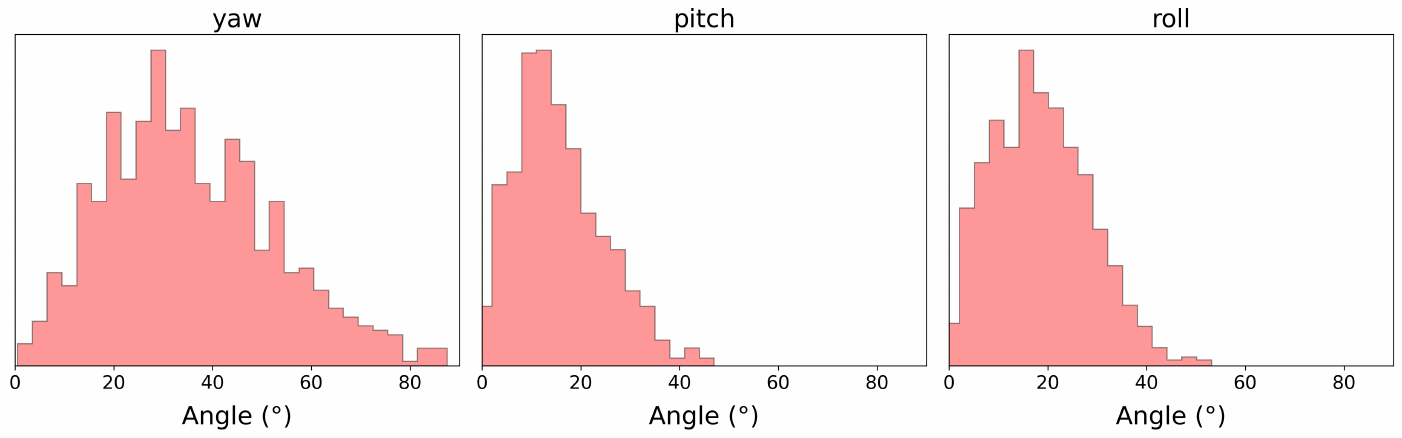}
            \caption{Distribution of the absolute pose differences for each Euler angle (yaw, pitch and roll) in our small benchmark dataset built from VoxCeleb1 test dataset.}
            \label{fig:hist}
        \end{figure*}

        We build a small benchmark in order to evaluate our method on challenging cases where the source and target faces have large head pose differences. Specifically, considering the VoxCeleb1~\cite{Nagrani17} test dataset, we pick 1000 pairs of images with large head pose distance (measured as the average of the absolute differences between the 3 Euler angles). In Fig.~\ref{fig:hist} we show the distribution of the absolute pose differences for each Euler angle, namely yaw, pitch and roll. This benchmark allows us to obtain deeper insights on the behavior of reenactment methods on challenging conditions. 
    
\begin{figure}[t]
            \centering
            \includegraphics[width=1.0\linewidth]{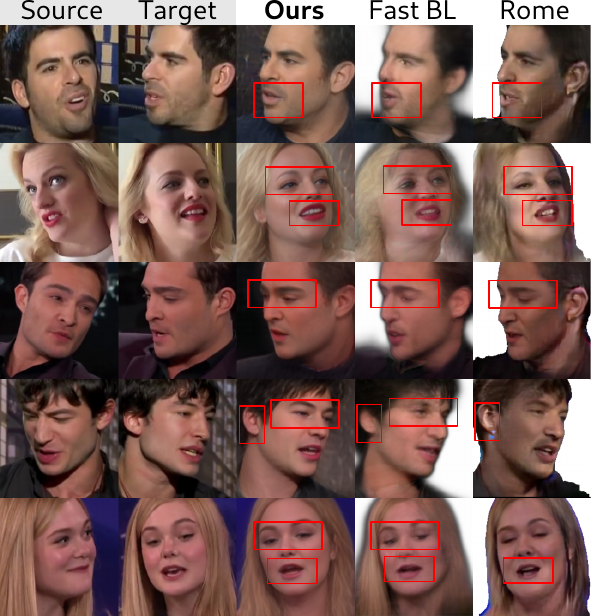}
            \caption{Qualitative results on the small benchmark with large head pose differences between the source and target faces.}
            \label{fig:artifact_details}
        \end{figure}

        In Fig.~\ref{fig:artifact_details}, we show comparisons of our method against the two state-of-the-art face reenactment methods, namely Fast BL~\cite{zakharov2020fast} and Rome~\cite{khakhulin2022rome}, on image pairs selected from the small benchmark described above. As shown, the source and target images have extreme head pose differences which makes it more challenging for face reenactment methods to generate realistic images. Nevertheless, our method is able to synthesize realistic faces even on extreme head poses. In Fig.~\ref{fig:artifact_details} we highlight (red boxes) details on the human faces, such as areas around the mouth and eyes,  where our method creates artifact free results, while Fast BL~\cite{zakharov2020fast} and Rome~\cite{khakhulin2022rome} generate blurry unrealistic images.  
    
    \subsection{Additional quantitative/qualitative results}~\label{sec:results}
    
\begin{figure*}[t]
            \centering
            \includegraphics[width=0.75\linewidth]{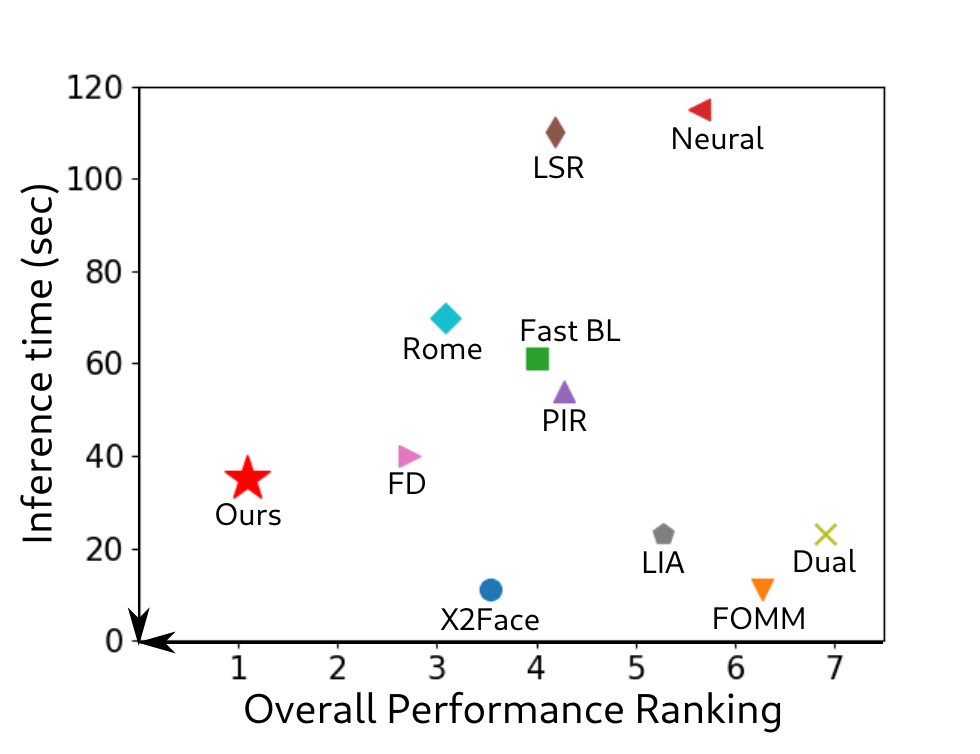}
            \caption{Comparisons in terms of the overall performance ranking of each method (presented in Table~\ref{table:reenactment_metrics_self_and_cross} of the main paper) and their inference time required to reenact a video of 200 frames. The arrows point towards the best results.}
            \label{fig:inference time}
        \end{figure*}

        In this Section, we present additional quantitative and qualitative results of the proposed method in comparison to state-of-the-art works. We compare all methods in terms of their inference time and their overall performance on the two tasks, namely, self and cross-subject reenactment. Specifically, in Table~\ref{table:inference_time_table}, we demonstrate the inference time of each method while reenacting a video of $200$ frames. To help drawing connections between the inference time and the performance of each method, we also report the performance ranking in Table~\ref{table:inference_time_table} (referred to as ``Perf. Rank''). Specifically, we consider the evaluation metrics that are reported in Table~\ref{table:reenactment_metrics_self_and_cross} of the main paper, and rank all methods with respect to each metric. Then, we average the ranking positions across all metrics on both self and cross-subject reenactment, for each method, to obtain its overall performance ranking. In Fig.~\ref{fig:inference time}, we present a plot of the two metrics (Inference time and Overall Performance Ranking). We note that our method achieves the best overall performance ranking, while also remaining competitive on the inference time. Additionally, while the inference time of X2Face~\cite{wiles2018x2face} and FOMM~\cite{siarohin2019first} is low compared to the other methods, as shown from the qualitative results they generate images with several visual artifacts.
    
\begin{figure*}[t]
            \centering
            \includegraphics[width=0.8\textwidth]{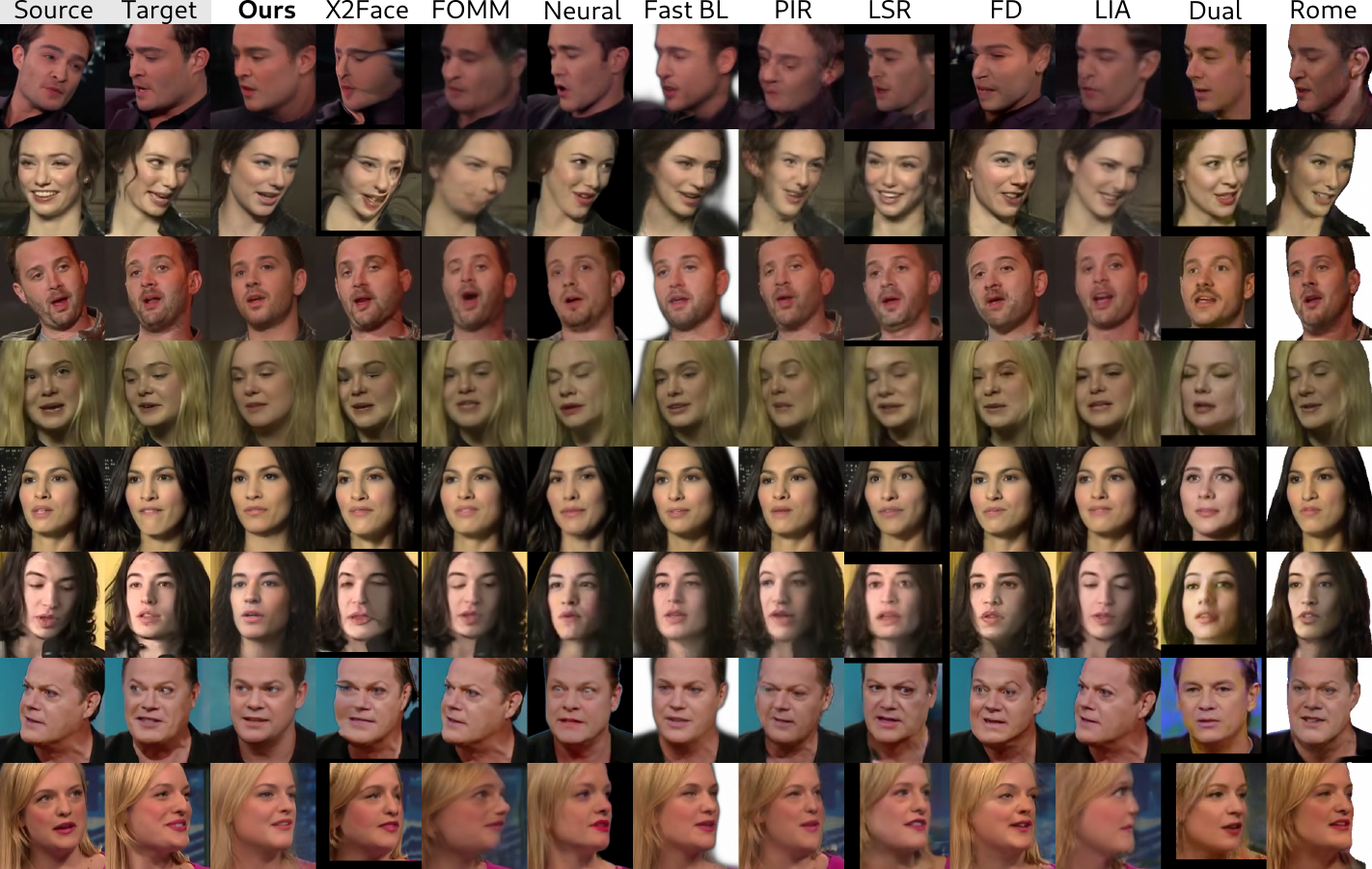}
            \caption{Additional qualitative results and comparisons on self-reenactment on VoxCeleb1 dataset~\cite{Nagrani17}. The first and second columns show the source and target faces.}
            \label{fig:self_vox1}
        \end{figure*}

\begin{figure*}[t]
        \centering
        \includegraphics[width=0.8\textwidth]{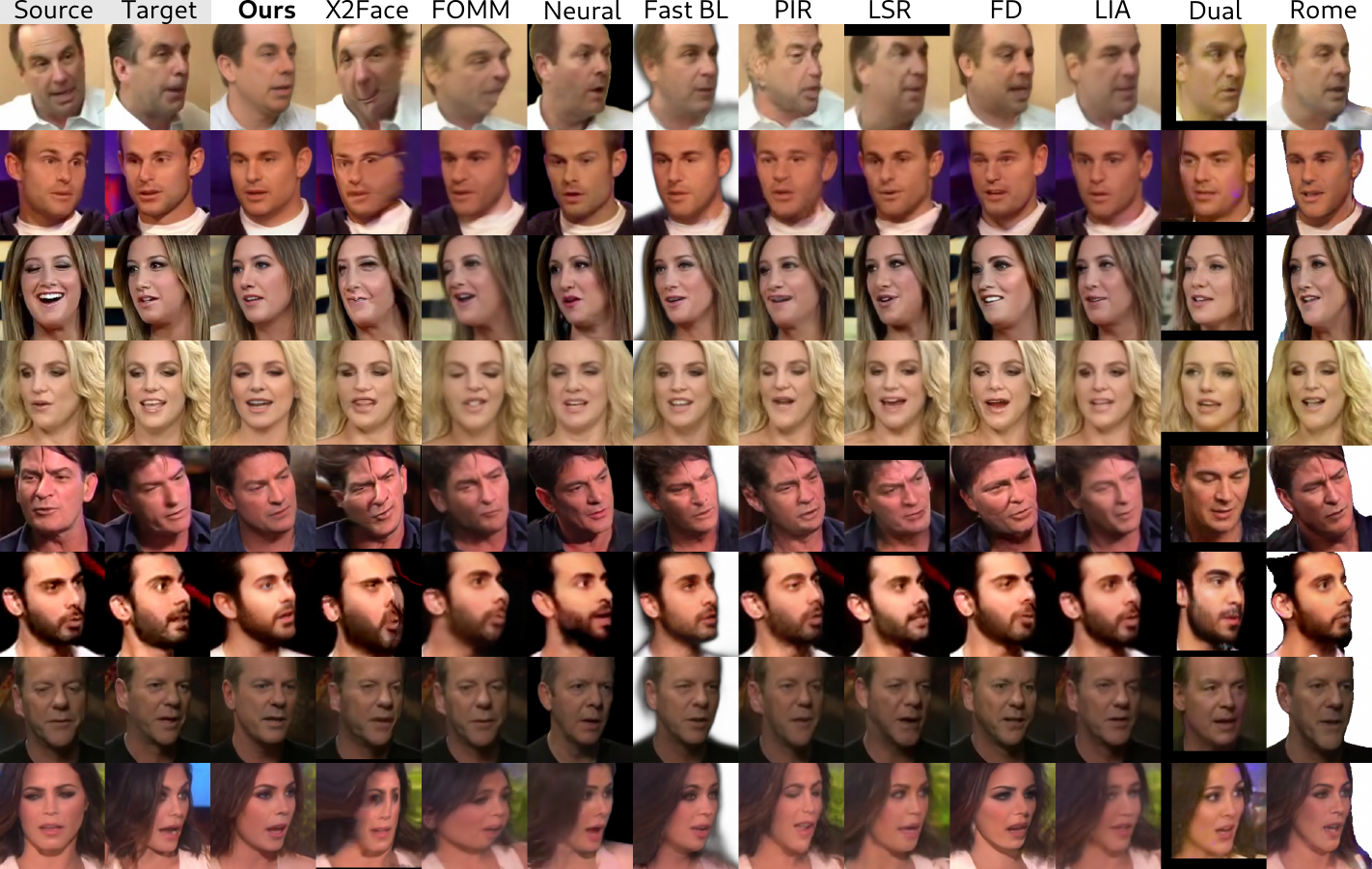}
        \caption{Qualitative results and comparisons on self-reenactment on VoxCeleb2 dataset~\cite{Chung18b}. The first and second columns show the source and target faces.}
        \label{fig:self_vox2}
        \end{figure*}

\begin{figure*}[t]
        \centering
        \includegraphics[width=0.8\textwidth]{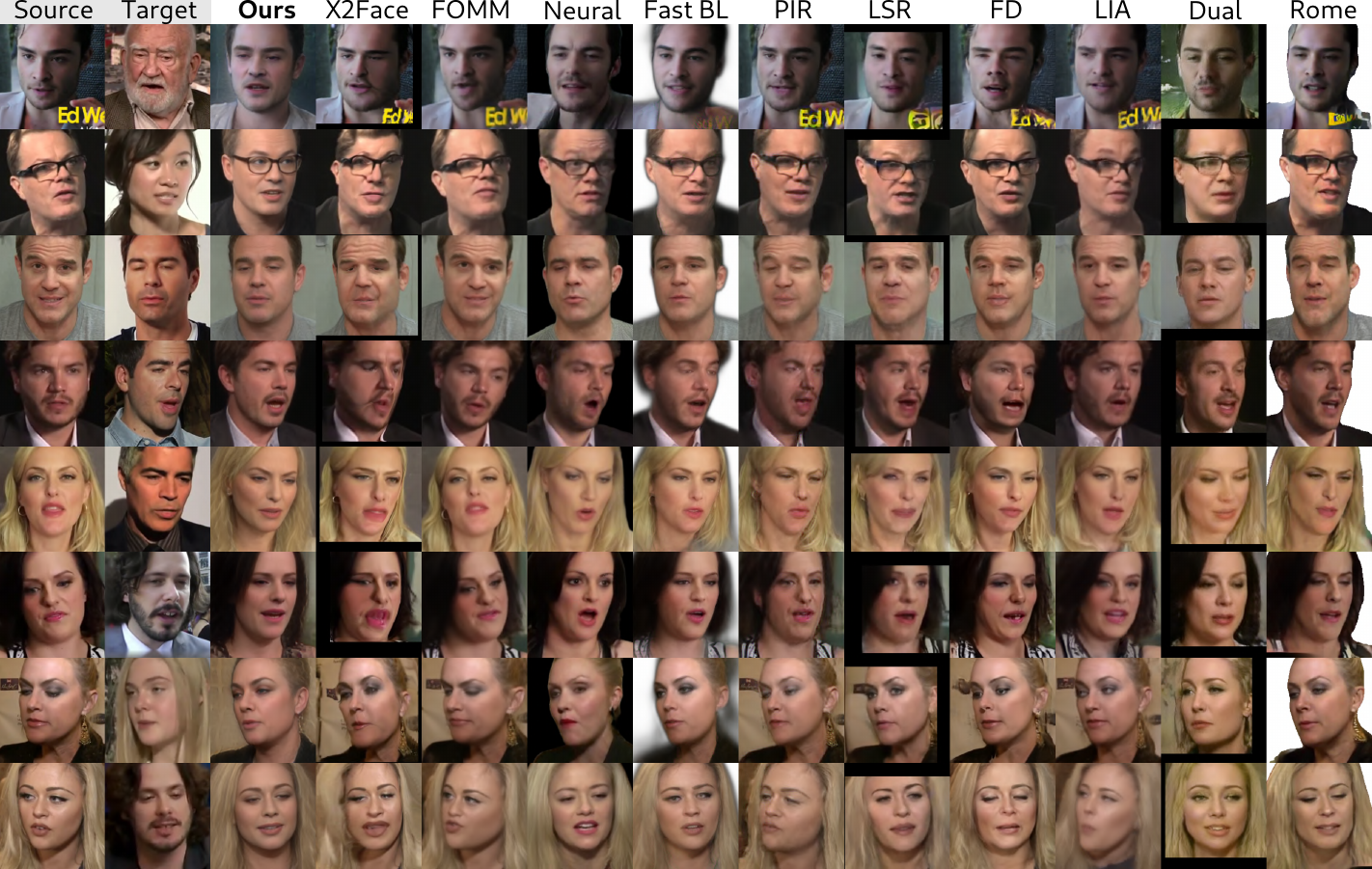}
        \caption{Additional qualitative results and comparisons on cross-subject reenactment on VoxCeleb1 dataset~\cite{Nagrani17}. The first and second columns show the source and target faces.}
        \label{fig:cross_vox1}
        \end{figure*}

\begin{figure*}[t]
            \centering
            \includegraphics[width=0.8\textwidth]{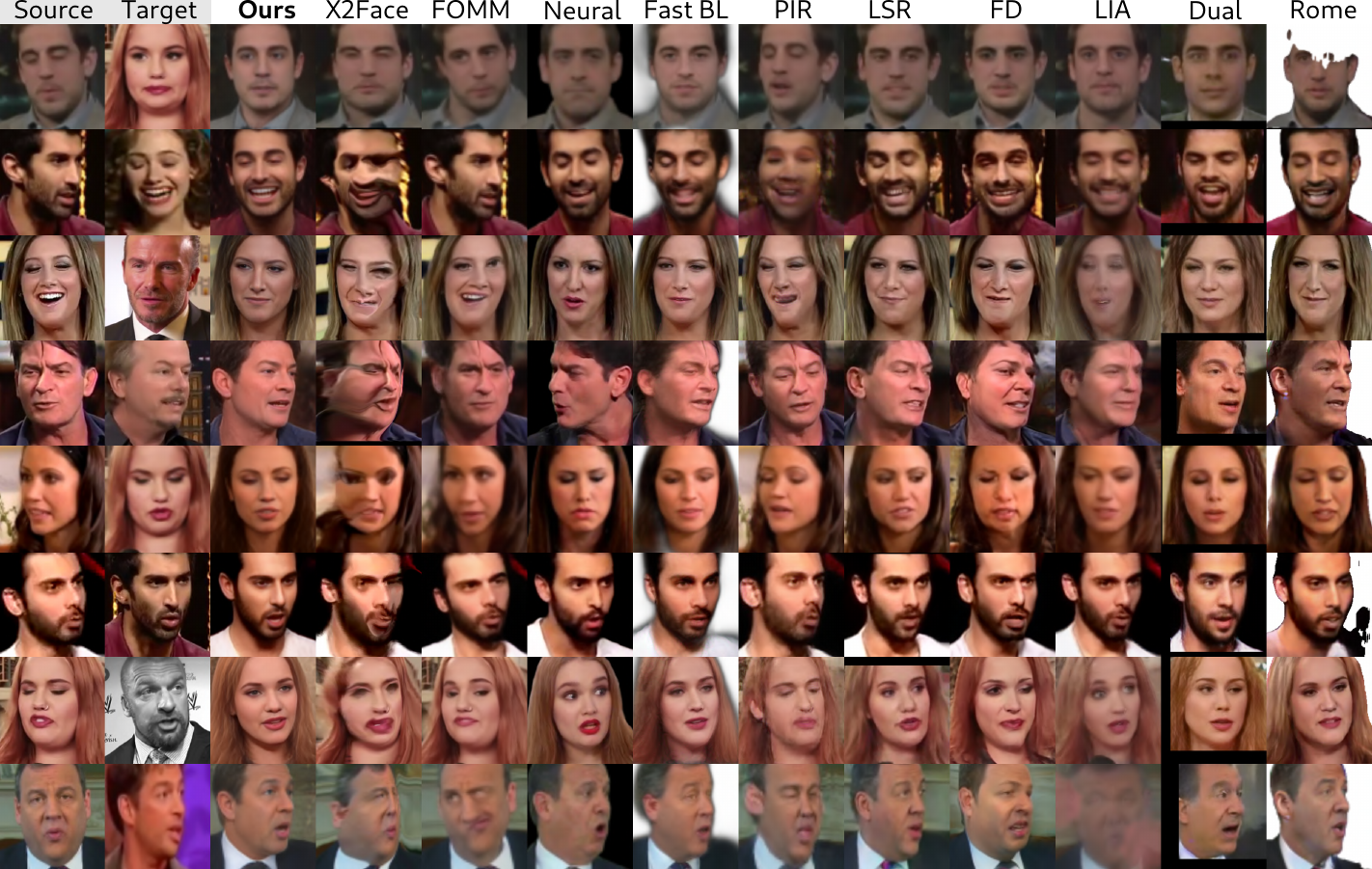}
            \caption{Qualitative results and comparisons on cross-subject reenactment on VoxCeleb2 dataset~\cite{Chung18b}. The first and second columns show the source and target faces, which are from different identities.}
            \label{fig:cross_vox2}
        \end{figure*}

\begin{figure*}[t]
            \centering
            \includegraphics[width=0.8\textwidth]{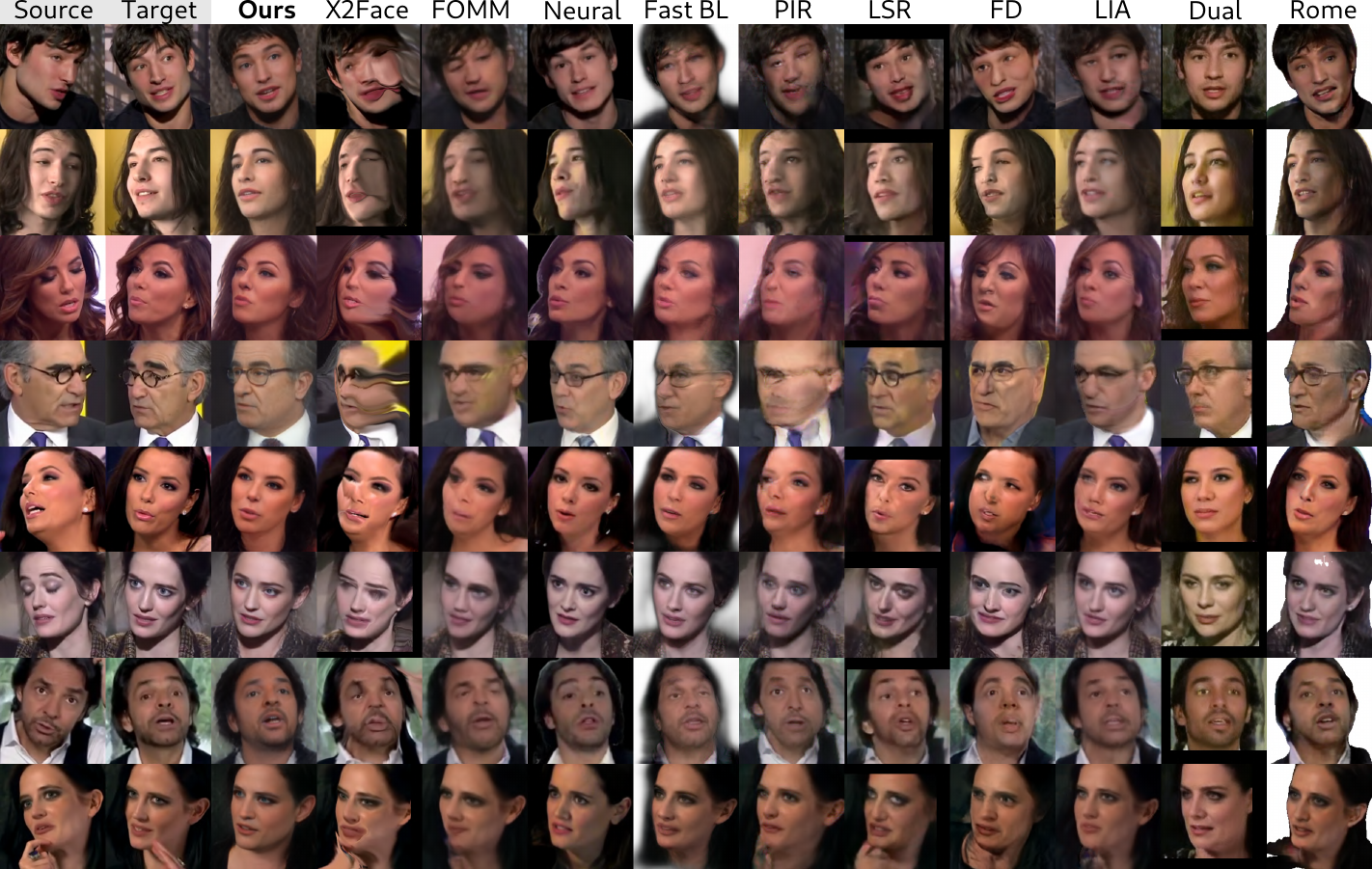}
            \caption{Additional qualitative results and comparisons on self reenactment using our small benchmark with image pairs that have large head pose differences. We show that our method presents robust results with artifact-free images, compared to the other state-of-the-art methods.}
            \label{fig:large_pose}
        \end{figure*}

        Moreover, we present further qualitative comparisons on the VoxCeleb1~\cite{Nagrani17}, as well as quantitative and qualitative results on VoxCeleb2~\cite{Chung18b}. Specifically, in Figs.~\ref{fig:self_vox1} and~\ref{fig:self_vox2} we show results on self reenactment, in Figs.~\ref{fig:cross_vox1} and~\ref{fig:cross_vox2} we show results on cross-subject reenactment, and in Fig.~\ref{fig:large_pose} we report results on self reenactment on the benchmark with extreme head pose differences described in Sect.~\ref{sec:extreme_pose}.
    
        Additionally, we quantitatively compare our method on the task of self reenactment on the VoxCeleb2 dataset~\cite{Chung18b} with the $10$ state-of-the-art methods, namely, X2Face~\cite{wiles2018x2face}, FOMM~\cite{siarohin2019first}, Neural~\cite{burkov2020neural}, Fast BL~\cite{zakharov2020fast}, PIR~\cite{ren2021pirenderer}, LSR~\cite{meshry2021learned}, FD~\cite{bounareli2022finding}, LIA~\cite{wang2021latent}, Dual~\cite{hsu2022dual}, and Rome~\cite{khakhulin2022rome}. In Table.~\ref{table:self_reenactment_metrics_vox2}, we present the quantitative results on self reenactment. As shown, our method outperforms all other methods both on identity preservation (CSIM) and on head pose (APD) and expression (AED) transfer (similarly to Section~\ref{ssec:comparisons}).
    
\begin{figure*}[t]
            \centering
            \includegraphics[width=1.0\textwidth]{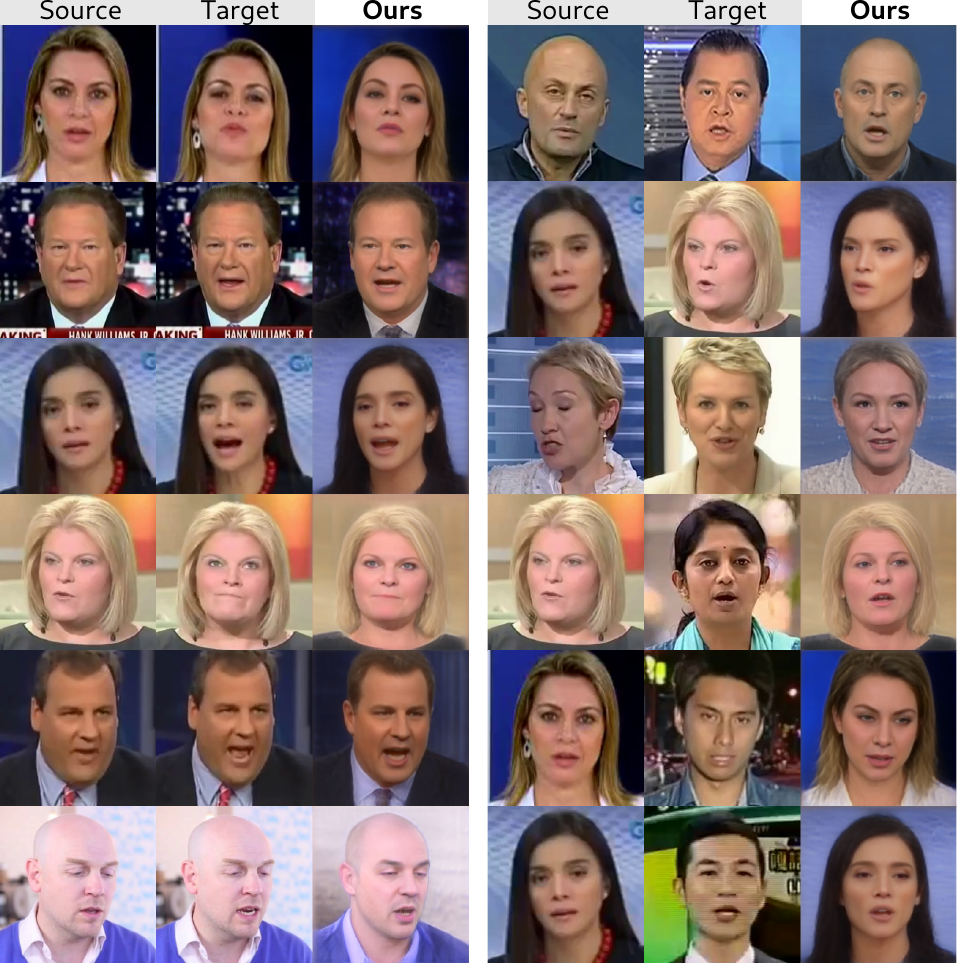}
            \caption{Additional qualitative results of our method on other video datasets such as FaceForensics~\cite{roessler2018faceforensics} and 300-VW~\cite{shen2015first}.}
            \label{fig:other_datasets}
        \end{figure*}

\begin{figure*}[t]
            \centering
            \includegraphics[width=1.0\textwidth]{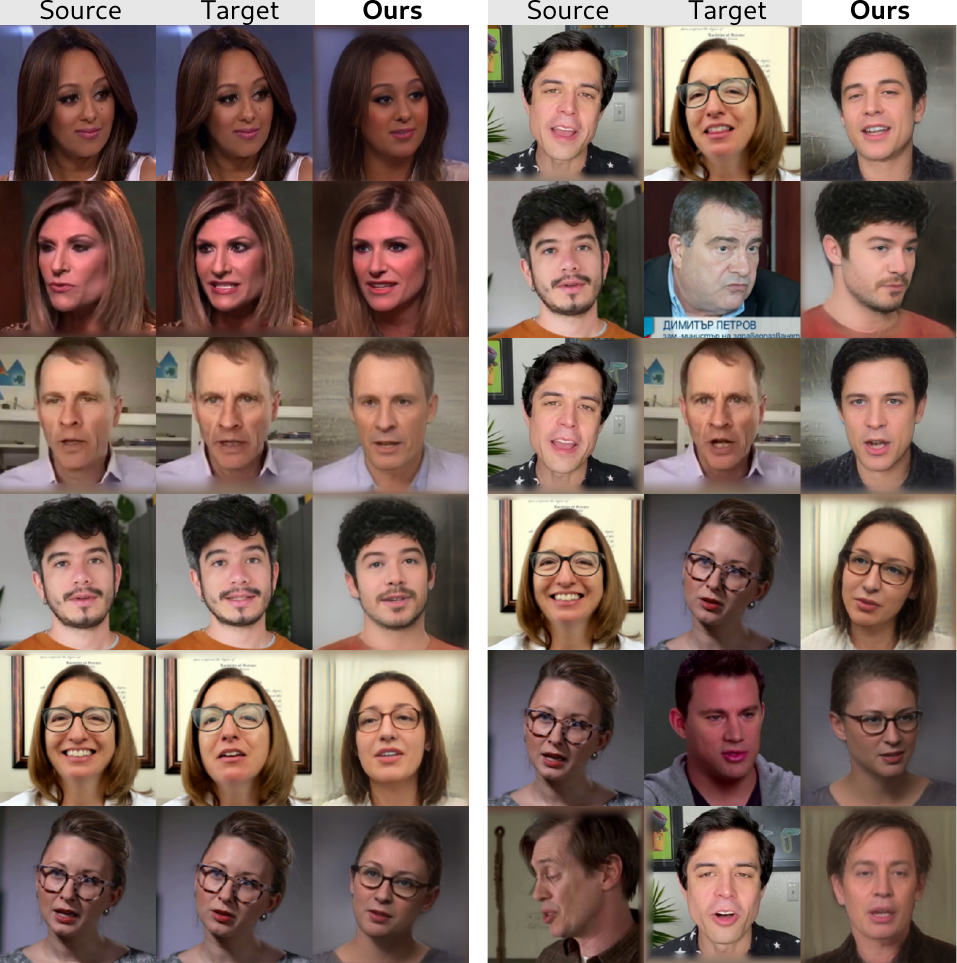}
            \caption{Additional qualitative results of our method on CelebV-HQ video dataset~\cite{zhu2022celebvhq}.}
            \label{fig:other_datasets_2}
        \end{figure*}

        Finally, in Figs.~\ref{fig:other_datasets},~\ref{fig:other_datasets_2}, we demonstrate results of our method, both on self and on cross-subject reenactment, on additional video datasets, namely, FaceForensics~\cite{roessler2018faceforensics}, 300-VW~\cite{shen2015first}, and CelebV-HQ~\cite{zhu2022celebvhq}, showing that the proposed method can generalise well on different video benchmarks.

    \subsection{Ethics consideration}\label{sec:ethics}
    
        Neural face reenactment methods allow for the creation of realistic talking head sequences that resemble the real ones. Consequently, besides being used for benevolent and useful purposes, such as in video conferencing, film and video production, arts, and education, we acknowledge that face reenactment methods, such as the proposed one, can also be misused towards malevolent purposes, such as deepfake fraud, that can harm individuals and can pose a greater societal threat.

        \begin{table}[t]
        \begin{center}
        \begin{tabular}{|l!{\vrule width 1.1pt}c|c|}
            \hline
            Method & Inf. time (sec)$\downarrow$  & Perf. Rank $\downarrow$ \\
            \noalign{\hrule height 1.2pt}
            X2Face~\cite{wiles2018x2face} & \textbf{11.0} & 3.5\\
            FOMM~\cite{siarohin2019first} & \textbf{11.0} & 6.3\\
            Neural~\cite{burkov2020neural} & 115.0 & 5.6\\
            Fast BL~\cite{zakharov2020fast} & 61.0 & 4.0\\  
            PIR~\cite{ren2021pirenderer} & 54.0 & 4.3\\
            LSR~\cite{meshry2021learned} & 110.0 & 4.2\\
            FD~\cite{bounareli2022finding} & 40.0 & \underline{2.7}\\
            LIA~\cite{wang2021latent} & \underline{23.0} & 5.3\\
            Dual~\cite{hsu2022dual} & \underline{23.0} & 6.9 \\
            Rome~\cite{khakhulin2022rome} & 70.0 & 3.1\\
            Ours & 37.0 & \textbf{1.1}\\
            \hline
        \end{tabular}
        \end{center}
        \caption{Quantitative comparisons on inference time and overall performance ranking (Perf. Rank) of all methods on self and cross-subject reenactment tasks (inference time is calculated while reenacting a video of 200 frames).}\label{table:inference_time_table}
        \end{table}

        \begin{table}[t]
        \begin{center}
        \begin{tabular}{|l!{\vrule width 1.1pt}c|c|c|}
            \hline
            Method & CSIM$\uparrow$  & APD$\downarrow$  & AED$\downarrow$ \\
            \noalign{\hrule height 1.2pt}
            X2Face~\cite{wiles2018x2face} & 0.60 & 2.4 & 10.6 \\\
            FOMM~\cite{siarohin2019first} & 0.57 & 5.1 & 13.6\\
            Neural~\cite{burkov2020neural} & 0.39 & 1.4 & 9.1\\
            Fast BL~\cite{zakharov2020fast} & 0.57 & 1.1 & 8.6\\  
            PIR~\cite{ren2021pirenderer} & 0.57 & 2.8 & 10.8\\
            LSR~\cite{meshry2021learned} & 0.61  & \underline{1.0} & 7.5\\
            FD~\cite{bounareli2022finding} & 0.59 & 1.3 & 7.3\\
            LIA~\cite{wang2021latent} & 0.64 &  2.5 & 8.7\\
            Dual~\cite{hsu2022dual} & 0.15 &  3.7 & 12.7\\
            Rome~\cite{khakhulin2022rome} & \underline{0.63} &  1.3 & \underline{5.9}\\
            Ours & \textbf{0.65} &  \textbf{ 0.5} & \textbf{5.2}\\
            \hline
        \end{tabular}
        \end{center}
        \caption{Quantitative results on the task of self reenactment on VoxCeleb2 dataset~\cite{Chung18b}.}\label{table:self_reenactment_metrics_vox2}
        \end{table}
        
\clearpage
\clearpage
{\small
\bibliographystyle{ieee_fullname}
\bibliography{egbib}

\begin{thebibliography}{10}\itemsep=-1pt

\bibitem{abdal2019image2stylegan}
Rameen Abdal, Yipeng Qin, and Peter Wonka.
\newblock Image2stylegan: How to embed images into the stylegan latent space?
\newblock In {\em Proceedings of the IEEE/CVF International Conference on
  Computer Vision}, pages 4432--4441, 2019.

\bibitem{abdal2021styleflow}
Rameen Abdal, Peihao Zhu, Niloy~J Mitra, and Peter Wonka.
\newblock Styleflow: Attribute-conditioned exploration of stylegan-generated
  images using conditional continuous normalizing flows.
\newblock {\em ACM Transactions on Graphics (ToG)}, 40(3), 2021.

\bibitem{alaluf2021restyle}
Yuval Alaluf, Or Patashnik, and Daniel Cohen-Or.
\newblock Restyle: A residual-based stylegan encoder via iterative refinement.
\newblock In {\em Proceedings of the IEEE/CVF International Conference on
  Computer Vision}, pages 6711--6720, 2021.

\bibitem{alaluf2022hyperstyle}
Yuval Alaluf, Omer Tov, Ron Mokady, Rinon Gal, and Amit Bermano.
\newblock Hyperstyle: Stylegan inversion with hypernetworks for real image
  editing.
\newblock In {\em Proceedings of the IEEE/CVF Conference on Computer Vision and
  Pattern Recognition}, pages 18511--18521, 2022.

\bibitem{bai2022high}
Qingyan Bai, Yinghao Xu, Jiapeng Zhu, Weihao Xia, Yujiu Yang, and Yujun Shen.
\newblock High-fidelity gan inversion with padding space.
\newblock In {\em European Conference on Computer Vision}, pages 36--53.
  Springer, 2022.

\bibitem{barattin2023attribute}
Simone Barattin, Christos Tzelepis, Ioannis Patras, and Nicu Sebe.
\newblock Attribute-preserving face dataset anonymization via latent code
  optimization.
\newblock In {\em Proceedings of the IEEE/CVF Conference on Computer Vision and
  Pattern Recognition}, pages 8001--8010, 2023.

\bibitem{bengio2009curriculum}
Yoshua Bengio, J{\'e}r{\^o}me Louradour, Ronan Collobert, and Jason Weston.
\newblock Curriculum learning.
\newblock In {\em Proceedings of the 26th annual international conference on
  machine learning}, pages 41--48, 2009.

\bibitem{blanz1999morphable}
Volker Blanz and Thomas Vetter.
\newblock A morphable model for the synthesis of 3d faces.
\newblock In {\em Proceedings of the 26th annual conference on Computer
  graphics and interactive techniques}, 1999.

\bibitem{bounareli2022finding}
Stella Bounareli, Vasileios Argyriou, and Georgios Tzimiropoulos.
\newblock Finding directions in gan's latent space for neural face reenactment.
\newblock {\em British Machine Vision Conference (BMVC)}, 2022.

\bibitem{bounareli2022StyleMask}
Stella Bounareli, Christos Tzelepis, Vasileios Argyriou, Ioannis Patras, and
  Georgios Tzimiropoulos.
\newblock Stylemask: Disentangling the style space of stylegan2 for neural face
  reenactment.
\newblock {\em IEEE Conference on Automatic Face and Gesture Recognition},
  2023.

\bibitem{burkov2020neural}
Egor Burkov, Igor Pasechnik, Artur Grigorev, and Victor Lempitsky.
\newblock Neural head reenactment with latent pose descriptors.
\newblock In {\em CVPR}, 2020.

\bibitem{Chung18b}
J.~S. Chung, A. Nagrani, and A. Zisserman.
\newblock Voxceleb2: Deep speaker recognition.
\newblock In {\em INTERSPEECH}, 2018.

\bibitem{deng2019arcface}
Jiankang Deng, Jia Guo, Niannan Xue, and Stefanos Zafeiriou.
\newblock Arcface: Additive angular margin loss for deep face recognition.
\newblock In {\em Proceedings of the IEEE/CVF conference on computer vision and
  pattern recognition}, pages 4690--4699, 2019.

\bibitem{dinh2022hyperinverter}
Tan~M Dinh, Anh~Tuan Tran, Rang Nguyen, and Binh-Son Hua.
\newblock Hyperinverter: Improving stylegan inversion via hypernetwork.
\newblock In {\em Proceedings of the IEEE/CVF Conference on Computer Vision and
  Pattern Recognition}, pages 11389--11398, 2022.

\bibitem{doukas2021headgan}
Michail~Christos Doukas, Stefanos Zafeiriou, and Viktoriia Sharmanska.
\newblock Headgan: One-shot neural head synthesis and editing.
\newblock In {\em Proceedings of the IEEE/CVF International Conference on
  Computer Vision}, pages 14398--14407, 2021.

\bibitem{feng2020deca}
Yao Feng, Haiwen Feng, Michael~J Black, and Timo Bolkart.
\newblock Learning an animatable detailed 3d face model from in-the-wild
  images.
\newblock {\em ACM Transactions on Graphics (TOG)}, 40(4):1--13, 2021.

\bibitem{ghosh2020gif}
Partha Ghosh, Pravir~Singh Gupta, Roy Uziel, Anurag Ranjan, Michael~J Black,
  and Timo Bolkart.
\newblock Gif: Generative interpretable faces.
\newblock In {\em 2020 International Conference on 3D Vision (3DV)}, pages
  868--878. IEEE, 2020.

\bibitem{hypernets}
David Ha, Andrew~M. Dai, and Quoc~V. Le.
\newblock Hypernetworks.
\newblock In {\em 5th International Conference on Learning Representations,
  {ICLR} 2017, Toulon, France, April 24-26, 2017, Conference Track
  Proceedings}, 2017.

\bibitem{ha2020marionette}
Sungjoo Ha, Martin Kersner, Beomsu Kim, Seokjun Seo, and Dongyoung Kim.
\newblock Marionette: Few-shot face reenactment preserving identity of unseen
  targets.
\newblock In {\em Proceedings of the AAAI conference on artificial
  intelligence}, volume~34, pages 10893--10900, 2020.

\bibitem{heusel2017gans}
Martin Heusel, Hubert Ramsauer, Thomas Unterthiner, Bernhard Nessler, and Sepp
  Hochreiter.
\newblock Gans trained by a two time-scale update rule converge to a local nash
  equilibrium.
\newblock {\em Advances in neural information processing systems}, 30, 2017.

\bibitem{hsu2022dual}
Gee-Sern Hsu, Chun-Hung Tsai, and Hung-Yi Wu.
\newblock Dual-generator face reenactment.
\newblock In {\em Proceedings of the IEEE/CVF Conference on Computer Vision and
  Pattern Recognition}, pages 642--650, 2022.

\bibitem{huang2020learning}
Po-Hsiang Huang, Fu-En Yang, and Yu-Chiang~Frank Wang.
\newblock Learning identity-invariant motion representations for cross-id face
  reenactment.
\newblock In {\em Proceedings of the IEEE/CVF Conference on Computer Vision and
  Pattern Recognition}, pages 7084--7092, 2020.

\bibitem{johnson2016perceptual}
Justin Johnson, Alexandre Alahi, and Li Fei-Fei.
\newblock Perceptual losses for real-time style transfer and super-resolution.
\newblock In {\em European conference on computer vision}, pages 694--711.
  Springer, 2016.

\bibitem{karras2019style}
Tero Karras, Samuli Laine, and Timo Aila.
\newblock A style-based generator architecture for generative adversarial
  networks.
\newblock In {\em Proceedings of the IEEE/CVF Conference on Computer Vision and
  Pattern Recognition}, pages 4401--4410, 2019.

\bibitem{karras2020analyzing}
Tero Karras, Samuli Laine, Miika Aittala, Janne Hellsten, Jaakko Lehtinen, and
  Timo Aila.
\newblock Analyzing and improving the image quality of stylegan.
\newblock In {\em Proceedings of the IEEE/CVF Conference on Computer Vision and
  Pattern Recognition}, pages 8110--8119, 2020.

\bibitem{khakhulin2022rome}
Taras Khakhulin, Vanessa Sklyarova, Victor Lempitsky, and Egor Zakharov.
\newblock Realistic one-shot mesh-based head avatars.
\newblock In {\em European Conference on Computer Vision}, pages 345--362.
  Springer, 2022.

\bibitem{kingma2014adam}
Diederik~P. Kingma and Jimmy Ba.
\newblock Adam: {A} method for stochastic optimization.
\newblock In Yoshua Bengio and Yann LeCun, editors, {\em 3rd International
  Conference on Learning Representations, {ICLR} 2015, San Diego, CA, USA, May
  7-9, 2015, Conference Track Proceedings}, 2015.

\bibitem{koujan2020head2head}
Mohammad~Rami Koujan, Michail~Christos Doukas, Anastasios Roussos, and Stefanos
  Zafeiriou.
\newblock Head2head: Video-based neural head synthesis.
\newblock In {\em 2020 15th IEEE International Conference on Automatic Face and
  Gesture Recognition (FG 2020)}, pages 16--23. IEEE, 2020.

\bibitem{meshry2021learned}
Moustafa Meshry, Saksham Suri, Larry~S Davis, and Abhinav Shrivastava.
\newblock Learned spatial representations for few-shot talking-head synthesis.
\newblock In {\em Proceedings of the IEEE/CVF International Conference on
  Computer Vision}, pages 13829--13838, 2021.

\bibitem{Nagrani17}
A. Nagrani, J.~S. Chung, and A. Zisserman.
\newblock Voxceleb: a large-scale speaker identification dataset.
\newblock In {\em INTERSPEECH}, 2017.

\bibitem{oldfield2021tensor}
James Oldfield, Markos Georgopoulos, Yannis Panagakis, Mihalis~A Nicolaou, and
  Ioannis Patras.
\newblock Tensor component analysis for interpreting the latent space of gans.
\newblock {\em British Machine Vision Conference (BMVC)}, 2021.

\bibitem{oldfield2023panda}
James Oldfield, Christos Tzelepis, Yannis Panagakis, Mihalis~A Nicolaou, and
  Ioannis Patras.
\newblock Panda: Unsupervised learning of parts and appearances in the feature
  maps of gans.
\newblock {\em International Conference on Learning Representations (ICLR)},
  2023.

\bibitem{park2019semantic}
Taesung Park, Ming-Yu Liu, Ting-Chun Wang, and Jun-Yan Zhu.
\newblock Semantic image synthesis with spatially-adaptive normalization.
\newblock In {\em Proceedings of the IEEE/CVF conference on computer vision and
  pattern recognition}, pages 2337--2346, 2019.

\bibitem{paszke2017automatic}
Adam Paszke, Sam Gross, Francisco Massa, Adam Lerer, James Bradbury, Gregory
  Chanan, Trevor Killeen, Zeming Lin, Natalia Gimelshein, Luca Antiga, et~al.
\newblock Pytorch: An imperative style, high-performance deep learning library.
\newblock {\em Advances in neural information processing systems},
  32:8026--8037, 2019.

\bibitem{patashnik2021styleclip}
Or Patashnik, Zongze Wu, Eli Shechtman, Daniel Cohen-Or, and Dani Lischinski.
\newblock Styleclip: Text-driven manipulation of stylegan imagery.
\newblock In {\em Proceedings of the IEEE/CVF International Conference on
  Computer Vision}, 2021.

\bibitem{radford2021learning}
Alec Radford, Jong~Wook Kim, Chris Hallacy, Aditya Ramesh, Gabriel Goh,
  Sandhini Agarwal, Girish Sastry, Amanda Askell, Pamela Mishkin, Jack Clark,
  et~al.
\newblock Learning transferable visual models from natural language
  supervision.
\newblock In {\em International Conference on Machine Learning}, pages
  8748--8763. PMLR, 2021.

\bibitem{ren2021pirenderer}
Yurui Ren, Ge Li, Yuanqi Chen, Thomas~H Li, and Shan Liu.
\newblock Pirenderer: Controllable portrait image generation via semantic
  neural rendering.
\newblock In {\em Proceedings of the IEEE/CVF International Conference on
  Computer Vision}, pages 13759--13768, 2021.

\bibitem{richardson2021encoding}
Elad Richardson, Yuval Alaluf, Or Patashnik, Yotam Nitzan, Yaniv Azar, Stav
  Shapiro, and Daniel Cohen-Or.
\newblock Encoding in style: a stylegan encoder for image-to-image translation.
\newblock In {\em Proceedings of the IEEE/CVF Conference on Computer Vision and
  Pattern Recognition}, pages 2287--2296, 2021.

\bibitem{roich2021pivotal}
Daniel Roich, Ron Mokady, Amit~H Bermano, and Daniel Cohen-Or.
\newblock Pivotal tuning for latent-based editing of real images.
\newblock {\em arXiv preprint arXiv:2106.05744}, 2021.

\bibitem{roessler2018faceforensics}
Andreas R\"ossler, Davide Cozzolino, Luisa Verdoliva, Christian Riess, Justus
  Thies, and Matthias Nie{\ss}ner.
\newblock Face{F}orensics: A large-scale video dataset for forgery detection in
  human faces.
\newblock {\em arXiv}, 2018.

\bibitem{shen2015first}
Jie Shen, Stefanos Zafeiriou, Grigoris~G Chrysos, Jean Kossaifi, Georgios
  Tzimiropoulos, and Maja Pantic.
\newblock The first facial landmark tracking in-the-wild challenge: Benchmark
  and results.
\newblock In {\em Proceedings of the IEEE international conference on computer
  vision workshops}, pages 50--58, 2015.

\bibitem{shen2020interfacegan}
Yujun Shen, Ceyuan Yang, Xiaoou Tang, and Bolei Zhou.
\newblock Interfacegan: Interpreting the disentangled face representation
  learned by gans.
\newblock {\em IEEE transactions on pattern analysis and machine intelligence},
  2020.

\bibitem{siarohin2019first}
Aliaksandr Siarohin, St{\'e}phane Lathuili{\`e}re, Sergey Tulyakov, Elisa
  Ricci, and Nicu Sebe.
\newblock First order motion model for image animation.
\newblock {\em Advances in Neural Information Processing Systems},
  32:7137--7147, 2019.

\bibitem{skorokhodov2022stylegan}
Ivan Skorokhodov, Sergey Tulyakov, and Mohamed Elhoseiny.
\newblock Stylegan-v: A continuous video generator with the price, image
  quality and perks of stylegan2.
\newblock In {\em Proceedings of the IEEE/CVF Conference on Computer Vision and
  Pattern Recognition}, pages 3626--3636, 2022.

\bibitem{vsubrtova2022chunkygan}
Ad{\'e}la {\v{S}}ubrtov{\'a}, David Futschik, Jan {\v{C}}ech, Michal
  Luk{\'a}{\v{c}}, Eli Shechtman, and Daniel S{\`y}kora.
\newblock Chunkygan: Real image inversion via segments.
\newblock In {\em European Conference on Computer Vision}, pages 189--204.
  Springer, 2022.

\bibitem{tewari2020stylerig}
Ayush Tewari, Mohamed Elgharib, Gaurav Bharaj, Florian Bernard, Hans-Peter
  Seidel, Patrick P{\'e}rez, Michael Zollhofer, and Christian Theobalt.
\newblock Stylerig: Rigging stylegan for 3d control over portrait images.
\newblock In {\em Proceedings of the IEEE/CVF Conference on Computer Vision and
  Pattern Recognition}, pages 6142--6151, 2020.

\bibitem{tov2021designing}
Omer Tov, Yuval Alaluf, Yotam Nitzan, Or Patashnik, and Daniel Cohen-Or.
\newblock Designing an encoder for stylegan image manipulation.
\newblock {\em ACM Transactions on Graphics (TOG)}, 40(4):1--14, 2021.

\bibitem{tzelepis2022contraclip}
Christos Tzelepis, James Oldfield, Georgios Tzimiropoulos, and Ioannis Patras.
\newblock Contraclip: Interpretable gan generation driven by pairs of
  contrasting sentences.
\newblock {\em arXiv preprint arXiv:2206.02104}, 2022.

\bibitem{tzelepis2021warpedganspace}
Christos Tzelepis, Georgios Tzimiropoulos, and Ioannis Patras.
\newblock Warpedganspace: Finding non-linear rbf paths in gan latent space.
\newblock In {\em Proceedings of the IEEE/CVF International Conference on
  Computer Vision}, pages 6393--6402, 2021.

\bibitem{unterthiner2018towards}
Thomas Unterthiner, Sjoerd van Steenkiste, Karol Kurach, Raphael Marinier,
  Marcin Michalski, and Sylvain Gelly.
\newblock Towards accurate generative models of video: A new metric \&
  challenges.
\newblock {\em arXiv preprint arXiv:1812.01717}, 2018.

\bibitem{voynov2020unsupervised}
Andrey Voynov and Artem Babenko.
\newblock Unsupervised discovery of interpretable directions in the gan latent
  space.
\newblock In {\em International Conference on Machine Learning}. PMLR, 2020.

\bibitem{wang2022high}
Tengfei Wang, Yong Zhang, Yanbo Fan, Jue Wang, and Qifeng Chen.
\newblock High-fidelity gan inversion for image attribute editing.
\newblock In {\em Proceedings of the IEEE/CVF Conference on Computer Vision and
  Pattern Recognition}, pages 11379--11388, 2022.

\bibitem{wang2021one}
Ting-Chun Wang, Arun Mallya, and Ming-Yu Liu.
\newblock One-shot free-view neural talking-head synthesis for video
  conferencing.
\newblock In {\em Proceedings of the IEEE/CVF conference on computer vision and
  pattern recognition}, pages 10039--10049, 2021.

\bibitem{wang2021latent}
Yaohui Wang, Di Yang, Francois Bremond, and Antitza Dantcheva.
\newblock Latent image animator: Learning to animate images via latent space
  navigation.
\newblock In {\em International Conference on Learning Representations}, 2021.

\bibitem{wiles2018x2face}
Olivia Wiles, A Koepke, and Andrew Zisserman.
\newblock X2face: A network for controlling face generation using images,
  audio, and pose codes.
\newblock In {\em Proceedings of the European conference on computer vision
  (ECCV)}, pages 670--686, 2018.

\bibitem{wu2021stylespace}
Zongze Wu, Dani Lischinski, and Eli Shechtman.
\newblock Stylespace analysis: Disentangled controls for stylegan image
  generation.
\newblock In {\em Proceedings of the IEEE/CVF Conference on Computer Vision and
  Pattern Recognition}, 2021.

\bibitem{xia2022gan}
Weihao Xia, Yulun Zhang, Yujiu Yang, Jing-Hao Xue, Bolei Zhou, and Ming-Hsuan
  Yang.
\newblock Gan inversion: A survey.
\newblock {\em IEEE Transactions on Pattern Analysis and Machine Intelligence},
  2022.

\bibitem{xu2022designing}
Chao Xu, Jiangning Zhang, Yue Han, Guanzhong Tian, Xianfang Zeng, Ying Tai,
  Yabiao Wang, Chengjie Wang, and Yong Liu.
\newblock Designing one unified framework for high-fidelity face reenactment
  and swapping.
\newblock In {\em European Conference on Computer Vision}, pages 54--71.
  Springer, 2022.

\bibitem{yang2021discovering}
Huiting Yang, Liangyu Chai, Qiang Wen, Shuang Zhao, Zixun Sun, and Shengfeng
  He.
\newblock Discovering interpretable latent space directions of gans beyond
  binary attributes.
\newblock In {\em Proceedings of the IEEE/CVF Conference on Computer Vision and
  Pattern Recognition}, pages 12177--12185, 2021.

\bibitem{Face2Face}
Kewei Yang, Kang Chen, Daoliang Guo, Song{-}Hai Zhang, Yuanchen Guo, and
  Weidong Zhang.
\newblock Face2face\({}^{\mbox{{\(\rho\)}}}\): Real-time high-resolution
  one-shot face reenactment.
\newblock In {\em Computer Vision - {ECCV} 2022 - 17th European Conference, Tel
  Aviv, Israel, October 23-27, 2022, Proceedings, Part {XIII}}. Springer, 2022.

\bibitem{yang2023just}
Qi Yang, Christos Tzelepis, Sergey Nikolenko, Ioannis Patras, and Aleksandr
  Farseev.
\newblock " just to see you smile": Smiley, a voice-guided guy gan.
\newblock In {\em Proceedings of the Sixteenth ACM International Conference on
  Web Search and Data Mining}, pages 1196--1199, 2023.

\bibitem{yin2022styleheat}
Fei Yin, Yong Zhang, Xiaodong Cun, Mingdeng Cao, Yanbo Fan, Xuan Wang, Qingyan
  Bai, Baoyuan Wu, Jue Wang, and Yujiu Yang.
\newblock Styleheat: One-shot high-resolution editable talking face generation
  via pretrained stylegan.
\newblock {\em arXiv preprint arXiv:2203.04036}, 2022.

\bibitem{zakharov2020fast}
Egor Zakharov, Aleksei Ivakhnenko, Aliaksandra Shysheya, and Victor Lempitsky.
\newblock Fast bi-layer neural synthesis of one-shot realistic head avatars.
\newblock In {\em ECCV}, 2020.

\bibitem{zakharov2019few}
Egor Zakharov, Aliaksandra Shysheya, Egor Burkov, and Victor Lempitsky.
\newblock Few-shot adversarial learning of realistic neural talking head
  models.
\newblock In {\em Proceedings of the IEEE/CVF International Conference on
  Computer Vision}, pages 9459--9468, 2019.

\bibitem{zhang2020freenet}
Jiangning Zhang, Xianfang Zeng, Mengmeng Wang, Yusu Pan, Liang Liu, Yong Liu,
  Yu Ding, and Changjie Fan.
\newblock Freenet: Multi-identity face reenactment.
\newblock In {\em Proceedings of the IEEE/CVF conference on computer vision and
  pattern recognition}, pages 5326--5335, 2020.

\bibitem{zhang2020eth}
Xucong Zhang, Seonwook Park, Thabo Beeler, Derek Bradley, Siyu Tang, and Otmar
  Hilliges.
\newblock Eth-xgaze: A large scale dataset for gaze estimation under extreme
  head pose and gaze variation.
\newblock In {\em European Conference on Computer Vision}, pages 365--381.
  Springer, 2020.

\bibitem{zhang2021flow}
Zhimeng Zhang, Lincheng Li, Yu Ding, and Changjie Fan.
\newblock Flow-guided one-shot talking face generation with a high-resolution
  audio-visual dataset.
\newblock In {\em Proceedings of the IEEE/CVF Conference on Computer Vision and
  Pattern Recognition}, pages 3661--3670, 2021.

\bibitem{zhu2022celebvhq}
Hao Zhu, Wayne Wu, Wentao Zhu, Liming Jiang, Siwei Tang, Li Zhang, Ziwei Liu,
  and Chen~Change Loy.
\newblock {CelebV-HQ}: A large-scale video facial attributes dataset.
\newblock In {\em ECCV}, 2022.

\end{thebibliography}
}

\end{document}